\documentclass{article}

\usepackage[final]{neurips_2022}

\author{%
  Vivien Cabannes\thanks{Work done while at INRIA / ENS / PSL. Contact the first author at \texttt{vivien.cabannes@gmail.com}.} \\
  Meta\\
  \And
  Francis Bach \\
  INRIA / ENS / PSL \\
  \And
  Vianney Perchet \\
  ENSAE \\
  \And
  Alessandro Rudi \\
  INRIA / ENS / PSL \\
}

\usepackage[T1]{fontenc}
\usepackage[utf8]{inputenc}
\usepackage[english]{babel}

\usepackage{newtxtext}
\usepackage{mathtools} % amsmath with fixes and additions
\usepackage{amsthm}
\allowdisplaybreaks    % allows page break over long align environment

\usepackage[nonewtxmathopt]{newtxmath} % This loads amsfonts and amssymb
\usepackage[cal=cm,bb=ams,frak=pxtx]{mathalpha}
\PassOptionsToPackage{hyphens}{url}
\usepackage[hyphens]{url}
\usepackage[hidelinks]{hyperref}

\usepackage{graphicx}
\usepackage{caption}
\usepackage{tikz}
\usetikzlibrary{hobby,decorations.markings}
\usepackage{pgfplots}
\pgfplotsset{compat=1.17}

\usepackage{booktabs}
\usepackage{microtype}
\usepackage{xcolor}
\usepackage{enumitem}
\setitemize{itemsep=0pt,parsep=0pt,topsep=0pt}
\setenumerate{itemsep=0pt,parsep=0pt,topsep=0pt}
\usepackage[ruled,vlined]{algorithm2e}

\usepackage{xfrac}

\usepackage{natbib}
\setcitestyle{round}
\newcommand{\card}[1]{\left\vert #1 \right\vert}
\newcommand{\diff}{\mathop{}\!\mathrm{d}}
\newcommand{\ind}[1]{\mathbf{1}_{#1}}
\newcommand{\prob}[1]{\Delta_{#1}}
\DeclareMathOperator{\E}{\mathbb{E}}
\DeclareMathOperator{\Pbb}{\mathbb{P}}
\DeclareMathOperator*{\argmax}{arg\,max}
\DeclareMathOperator*{\argmin}{arg\,min}
\newcommand{\uniform}[1]{{\cal U}\paren{#1}}

\DeclareMathOperator{\Span}{Span}

\DeclareMathOperator{\op}{op}
\DeclareMathOperator{\sign}{sign}

\DeclareMathOperator*{\supp}{supp}

\newcommand{\N}{\mathbb{N}}

\newcommand{\R}{\mathbb{R}}

\renewcommand{\brace}[1]{\left\{ #1 \right\}}
\newcommand{\bracket}[1]{\left[ #1 \right]}
\newcommand{\floor}[1]{\left\lfloor #1 \right\rfloor}

\newcommand{\paren}[1]{\left( #1 \right)}
\newcommand{\midvert}{\,\middle\vert\,}
\newcommand{\midvertvert}{\,\middle\vert\vert\,}

\newcommand{\abs}[1]{\left| #1 \right|}

\newcommand{\norm}[1]{\left\| #1 \right\|}
\newcommand{\scap}[2]{\left\langle #1, #2 \right\rangle}

\newcommand{\Sfrak}{\mathfrak{S}}
\newcommand{\X}{\mathcal{X}}
\newcommand{\Y}{\mathcal{Y}}

\renewcommand{\epsilon}{\varepsilon}
\renewcommand{\phi}{\varphi}

\theoremstyle{plain}
\newtheorem{theorem}{Theorem}

\newtheorem{lemma}{Lemma}
\newtheorem{proposition}[lemma]{Proposition}
\newtheorem{definition}[lemma]{Definition}
\newtheorem{assumption}{Assumption}
\newtheorem{remark}[lemma]{Remark}
\newtheorem{example}{Example}

\newcommand{\myfunction}[5]{
\begin{array}{cccc}
	#1 : & #2 & \rightarrow & #3 \\
	 & #4 & \rightarrow & #5
\end{array}
}

\definecolor{coolred}{RGB}{200, 60, 100}

\title{Active Labeling: Streaming Stochastic Gradients}

\begin{document}
\maketitle

\begin{abstract}
	The workhorse of machine learning is stochastic gradient descent.
To access stochastic gradients, it is common to consider iteratively input/output pairs of a training dataset.
Interestingly, it appears that one does not need full supervision to access stochastic gradients, which is the main motivation of this paper.
After formalizing the "active labeling" problem, which focuses on active learning with partial supervision, we provide a streaming technique that provably minimizes the ratio of generalization error over the number of samples.
We illustrate our technique in depth for robust regression.
\end{abstract}

\section{Introduction}

A large amount of the current hype around artificial intelligence was fueled by the recent successes of supervised learning.
Supervised learning consists in designing an algorithm that maps inputs to outputs by learning from a set of input/output examples.
When accessing many samples, and given enough computation power, this framework is able to tackle complex tasks.
Interestingly, many of the difficulties arising in practice do not emerge from choosing the right statistical model to solve the supervised learning problem, but from the problem of collecting and cleaning enough data \cite[see Chapters 1 and 2 of][for example]{Geron2017}.
Those difficulties are not disjoint from the current trends toward data privacy regulations \citep{GDPR}.
This fact motivates this work, where we focus on how to efficiently collect information to carry out the learning process.

In this paper, we formalize the ``active labeling'' problem for weak supervision, where the goal is to learn a target function by acquiring the most informative dataset given a restricted budget for annotation.
We focus explicitly on weak supervision that comes as a set of label candidates for each input, aiming to partially supervise input data in the most efficient way to guide a learning algorithm.
We also restrict our study to the streaming variant where, for each input, only a single partial information can be collected about its corresponding output.
The crux of this work is to leverage the fact that full supervision is not needed to acquire unbiased stochastic gradients, and perform stochastic gradient descent.

The following summarizes our contributions.
\begin{enumerate}
  \item First, we introduce the ``active labeling'' problem, which is a relevant theoretical framework that encompasses many useful problems encountered by practitioners trying to annotate their data in the most efficient fashion, as well as its streaming variation, in order to deal with privacy preserving issues. This is the focus of Section~\ref{sgd:sec:fram}.
  \item Then, in Section~\ref{sgd:sec:sgd}, we give a high-level framework to access unbiased stochastic gradients with weak information only. This provides a simple solution to the streaming ``active labeling'' problem.
  \item Finally, we detail this framework for a robust regression task in Section~\ref{sgd:sec:median}, and provide an algorithm whose optimality is proved in Section~\ref{sgd:sec:stat}.
\end{enumerate}
As a proof of concept, we provide numerical simulations in Section~\ref{sgd:sec:exp}.
We conclude with a high-level discussion around our methods in Section~\ref{sgd:sec:discussion}.

\paragraph{Related work.}
Active query of information is relevant to many settings.
The most straightforward applications are searching games, such as Bar Kokhba or twenty questions \citep{Walsorth1882}.
We refer to \citet{Pelc2002} for an in-depth survey of such games, especially when liars introduce uncertainty, and their relations with coding on noisy channels.
But applications are much more diverse, {\em e.g.} for numerical simulation \citep{Chevalier2014}, database search \citep{Qarabaqi2014}, or shape recognition \citep{Geman1993}, to name a few.

In terms of motivations, many streams of research can be related to this problem, such as experimental design \citep{Chernoff1959}, statistical queries \citep{Kearns1998,Fotakis2021}, crowdsourcing \citep{Doan2011}, or aggregation methods in weak supervision \citep{Ratner2020}.
More precisely, ``active labeling''\footnote{Note that the wording ``active labeling'' has been more or less used as synonymous of ``active learning'' \cite[\emph{e.g.,}][]{Wang2014}. In contrast, we use ``active labeling'' to design ``active weakly supervised learning''.} consists in having several inputs and querying partial information on the labels. It is close to active learning \citep{Settles2010,Dasgupta2011,Hanneke2014}, where there are several inputs, but exact outputs are queried; and to active ranking \citep{Valiant1975,Ailon2011,Braverman2019}, where partial information is queried, but there is only one input.
The streaming variant introduces privacy preserving constraints, a problem that is usually tackled through the notion of differential privacy \citep{Dwork2006}.

In terms of formalization, we build on the partial supervision formalization of \cite{Cabannes2020}, which casts weak supervision as sets of label candidates and generalizes semi-supervised learning \citep{Chapelle2006}.
Finally, our sequential setting with a unique final reward is similar to combinatorial bandits in a pure-exploration setting \citep{Garivier2016,Fiez2019}.

\section{The ``active labeling'' problem}
\label{sgd:sec:fram}

Supervised learning is traditionally modeled in the following manner.
Consider $\X$ an input space, $\Y$ an output space, $\ell:\Y\times\Y\to\R$ a loss function, and $\rho\in\prob{\X\times\Y}$ a joint probability distribution.
The goal is to recover the function
\begin{equation}
  \label{sgd:eq:obj}
  f^*\in\argmin_{f:\X\to\Y} {\cal R}(f) := \E_{(X, Y)\sim\rho}[\ell(f(X), Y)],
\end{equation}
yet, without accessing $\rho$, but a dataset of independent samples distributed according to $\rho$, ${\cal D}_n = (X_i, Y_i)_{i\leq n}\sim\rho^{\otimes n}$.
In practice, accessing data comes at a cost, and it is valuable to understand the cheapest way to collect a dataset allowing to discriminate $f^*$.

We shall suppose that the input data $(X_i)_{i\leq n}$ are easy to collect, yet that labeling those inputs to get outputs $(Y_i)_{i\leq n}$ demands a high amount of work.
For example, it is relatively easy to scrap the web or medical databases to access radiography images, but labeling them by asking radiologists to recognize tumors on zillions of radiographs will be both time-consuming and expensive.
As a consequence, \emph{we assume the $(X_i)_{i\leq n}$ given but the $(Y_i)_{i\leq n}$ unknown}.
As getting information on the labels comes at a cost ({\em e.g.}, paying a pool of label workers, or spending your own time), given a budget constraint, what information should we query on the labels?

To quantify this problem, we will assume that {\em we can sequentially and adaptively query $T$ information of the type $\ind{Y_{i_t} \in S_t}$, for any index $i_t\in\brace{1, \cdots, n}$ and any set of labels $S_t \subset \Y$ (belonging to a specified set of subsets of $\Y$).}
Here, $t \in \brace{1, \cdots, T}$ indexes the query sequence, and $T\in\N$ is a fixed budget.
The goal is to optimize the design of the sequence $(i_t, S_t)$ in order to get the best estimate of $f^*$ in terms of risk minimization~\eqref{sgd:eq:obj}.
In the following, we give some examples to make this setting more concrete.

\begin{example}[Classification with attributes]
  Suppose that a labeler is asked to provide fine-grained classes on images \citep{Krause2016,Zheng2019}, such as the label ``caracal'' in Figure~\ref{sgd:fig:caracal}.
  This would be difficult for many people. Yet, it is relatively easy to recognize that the image depicts a ``feline'' with ``tufted-ears'' and ``sandy color''.
  As such, a labeler can give the weak information that $Y$ belongs to the set ``feline'', $S_1 = \brace{\text{``cat'', ``lion'', ``tiger''}, \dots}$, and the set ``tufted ears'', $S_2 = \brace{\text{``Great horned owl'', ``Aruacana chicken''}, \dots}$.
  This is enough to recognize that $Y \in S_1 \cap S_2 = \brace{\text{``caracal''}}$.
  The question $\ind{Y\in S_1}$, corresponds to asking if the image depicts a feline.
  Literature on hierarchical classification and autonomic taxonomy construction provides interesting ideas for this problem \citep[{\em e.g.},][]{Cesa-Bianchi2006,Gangaputra2006}.
\end{example}

\begin{example}[Ranking with partial ordering]
  Consider a problem where for a given input $x$, characterizing a user, we are asked to deduce their preferences over $m$ items.
  Collecting such a label requires knowing the exact ordering of the $m$ items induced by a user. This might be hard to ask for.
  Instead, one can easily ask the user which items they prefer in a collection of a few items. The user's answer will give weak information about the labels, which can be modeled as knowing $\ind{Y_i\in S} = 1$, for $S$ the set of total orderings that satisfy this partial ordering.
  We refer the curious reader to active ranking and dueling bandits for additional contents \citep{Jamieson2011,Bengs2021}.
\end{example}

\begin{example}[Pricing a product]
  Suppose that we want to sell a product to a consumer characterized by some features $x$, this consumer is ready to pay a price $y\in \R$ for this product.
  We price it $f(x)\in\R$, and we observe $\ind{f(x) < y}$, that is if the consumer is willing to buy this product at this price tag or not \citep{Cesa-Bianchi2019,Liu2021}.
  Although, in this setting, the goal is often to minimize the regret, which contrasts with our pure exploration setting.
\end{example}

As a counter-example, our assumptions are not set to deal with missing data, {\em i.e.} if some coordinates of some input feature vectors $X_i$ are missing \citep{Rubin1976}.
Typically, this happens when input data comes from different sources ({\em e.g.}, when trying to predict economic growth from country information that is self-reported).

\paragraph{Streaming variation.}
The special case of the active labeling problem we shall consider consists in its variant without resampling.
This corresponds to the online setting where one can only ask one question by sample, formally $i_t=t$.
This setting is particularly appealing for privacy concerns, in settings where the labels $(Y_i)$ contain sensitive information that should not be revealed totally.
For example, some people might be more comfortable giving a range over a salary rather than the exact value; or in the context of polling, one might not call back a previous respondent characterized by some features $X_i$ to ask them again about their preferences captured by $Y_i$.
Similarly, the streaming setting is relevant for web marketing, where inputs model new users visiting a website, queries model sets of advertisements chosen by an advertising company, and one observes potential clicks.

\section{Weak information as stochastic gradients}
\label{sgd:sec:sgd}

In this section, we discuss how unbiased stochastic gradients can be accessed through weak information.

Suppose that we model $f = f_\theta$ for some Hilbert space $\Theta\ni\theta$.
With some abuse of notations, let us denote $\ell(x, y, \theta) := \ell(f_\theta(x), y)$.
We aim to minimize
\(
{\cal R}(\theta) = \E_{(X, Y)}\bracket{\ell(X, Y, \theta)}.
\)
Assume that ${\cal R}$ is differentiable (or sub-differentiable) and denote its gradients by $\nabla_\theta {\cal R}$.

\begin{definition}[Stochastic gradient]
  A stochastic gradient of ${\cal R}$ is any random function $G:\Theta \to \Theta$ such that
  \(
  \E[G(\theta)] = \nabla_\theta{\cal R}(\theta).
  \)
  Given some step size function $\gamma:\N\to\R^*$, a stochastic gradient descent (SGD) is a procedure, $(\theta_t) \in \Theta^\N$, initialized with some $\theta_0$ and updated as
  \(
  \theta_{t+1} = \theta_t - \gamma(t) G(\theta_t),
  \)
  where the realization of $G(\theta_t)$ given $\theta_t$ is independent of the previous realizations of $G(\theta_{s})$ given $\theta_s$.
\end{definition}

In supervised learning, SGD is usually performed with the stochastic gradients $\nabla_\theta \ell(X, Y, \theta)$.
More generally, stochastic gradients are given by
\begin{equation}
  \label{sgd:eq:sgd_def}
  G(\theta) = \ind{\nabla_\theta \ell(X, Y, \theta) \in T}\cdot \tau(T),
\end{equation}
for $\tau:{\cal T}\to\Theta$ with ${\cal T} \subset 2^\Theta$ a set of subsets of $\Theta$, and $T$ a random variable on ${\cal T}$, such that
\begin{equation}
  \label{sgd:eq:condition}
  \forall\, \theta \in\Theta,\quad \E_T[\ind{\theta\in T} \cdot \tau(T)] = \theta.
\end{equation}
Stated otherwise, if you have a way to image a vector $\theta$ from partial measurements $\ind{\theta\in T}$ such that you can reconstruct this vector in a linear fashion \eqref{sgd:eq:condition}, then it provides you a generic strategy to get an unbiased stochastic estimate of this vector from a partial measurement \eqref{sgd:eq:sgd_def}.

For $\psi:\Y\to\Theta$ a function from $\Y$ to $\Theta$ ({\em e.g.}, $\psi = \nabla_\theta(X, \cdot, \theta)$), a question $\ind{\psi(Y) \in T}$ translates into a question $\ind{Y\in S}$ for some set $S = \psi^{-1}(T) \subset \Y$, meaning that the stochastic gradient \eqref{sgd:eq:sgd_def} can be evaluated from a single query.
As a proof of concept, we derive a generic implementation for $T$ and $\tau$ in Appendix \ref{sgd:app:generic}.
This provides a generic SGD scheme to learn functions from weak queries when there are no constraints on the sets to query.

\begin{remark}[Cutting plane methods]
  While we provide here a descent method, one could also develop cutting-plane/ellipsoid methods to localize $\theta^*$ according to weak information, which corresponds to the techniques developed for pricing by \cite{Cohen2020} and related literature.
\end{remark}

\section{Median regression}
\label{sgd:sec:median}

In this section, we focus on efficiently acquiring weak information providing stochastic gradients for regression problems.
In particular, we motivate and detail our methods for the absolute deviation loss.

Motivated by seminal works on censored data \citep{Tobin1958}, we shall suppose that {\em we query half-spaces}.
For an output $y\in\Y=\R^m$, and any hyper-plane $z + u^\perp \subset \R^m$ for $z \in \R^m$, $u \in \mathbb{S}^{m-1}$, we can ask a labeler to tell us which half-space $y$ belongs to.
Formally, {\em we access the quantity $\sign(\scap{y - z}{u})$ for a given unit cost}.
Such an imaging scheme where one observes summations of its components rather than a vector itself bears similarity with compressed sensing.
To provide further illustration, this setting could help to price products while selling bundles: where the context $x$ characterizes some users, web-pages or/and advertisement companies; the label $y\in\R^m$ corresponds to the value associated to $m$ different items, such as stocks composing an index, or advertisement spots; and the observation $\sign(\scap{y}{u} - c)$ (with $c = \scap{z}{u}$) captures if the user $x$ buys the basket with weights $u\in\mathbb{S}^{m-1}$ when it is priced $c$.

\paragraph{Least-squares.}
For regression problems, it is common to look at the mean square loss
\[
  \ell(X, Y, \theta) = \norm{f_\theta(X) - Y}^2,\qquad
  \nabla_\theta\ell(X, Y, \theta) = 2(f_\theta(X) - Y)^\top Df_\theta(X),
\]
where $Df_\theta(x)\in\Y\otimes\Theta$ denotes the Jacobian of $\theta \to f_\theta(x)$.
In rich parametric models, it is preferable to ask questions on $Y\in\Y$ rather than on gradients in $\Theta$ which is a potentially much bigger space.
If we assume that $Y$ and $f_\theta(X)$ are bounded in $\ell^2$-norm by $M \in \R_+$, we can adapt~\eqref{sgd:eq:sgd_def} and \eqref{sgd:eq:condition} through the fact that for any $z \in \Y$, such that $\norm{z} \leq 2M$, as proven in Appendix \ref{sgd:app:generic},
\[
  \E_{U, V}\bracket{\ind{\scap{z}{U} \geq V}\cdot U}
  = c_1\cdot z,\quad\text{where}\quad
  c_1 = \E_{U, V}\bracket{\ind{\scap{e_1}{U} \geq V} \cdot\scap{e_1}{U}}
  = \frac{\pi^{3/2}}{2M (m^2 + 4m + 3)},
\]
for $U$ uniform on the sphere $\mathbb{S}^{m-1}$ and $V$ uniform on $[0, 2M]$.
Applied to $z = f_\theta(X) - Y$, it designs an SGD procedure by querying information of the type
\(
\ind{\scap{Y}{U} < \scap{f_\theta(X)}{U} - V}.
\)

\paragraph{A case for median regression.}
Motivated by robustness purposes, we will rather expand on median regression.
In general, we would like to learn a function that, given an input, replicates the output of I/O samples generated by the joint probability $\rho$.
In many instances, $X$ does not characterize all the sources of variations of $Y$, {\em i.e.} input features are not rich enough to characterize a unique output, leading to randomness in the conditional distributions $(Y \vert X)$.
When many targets can be linked to a vector $x\in\X$, how to define a consensual $f(x)$?
For analytical reasons, statisticians tend to use the least-squares error which corresponds to asking for $f(x)$ to be the mean of the distribution $\paren{Y\vert X=x}$.
Yet, means are known to be too sensitive to rare but large outputs \citep[see \emph{e.g.},][]{Huber1981}, and cannot be defined as good and robust consensus in a world of heavy-tailed distributions.
This contrasts with the median, which, as a consequence, is often much more valuable to summarize a range of values.
For instance, median income is preferred over mean income as a population indicator \citep[see \emph{e.g.},][]{USCensus}.

\paragraph{Median regression.} The geometric median is variationally defined through the absolute deviation loss, leading to
\begin{equation}
  \label{sgd:eq:median}
  \ell(X, Y, \theta) = \norm{f_\theta(X) - Y},\qquad
  \nabla_\theta \ell(X, Y, \theta) =
  \paren{\frac{f_\theta(X) - Y}{\norm{f_\theta(X) - Y}}}^\top
  Df_\theta(X).
\end{equation}
Similarly to the least-squares case, we can access weakly supervised stochastic gradients through the fact that for $z \in \mathbb{S}^{m-1}$, as shown in Appendix \ref{sgd:app:generic},
\begin{equation}
  \label{sgd:eq:median_sgd}
  \E_{U}\bracket{\sign\paren{\scap{z}{U}}\cdot U}
  = c_2\cdot z,\quad\text{where}\quad
  c_2 = \E_{U}\bracket{\sign\paren{\scap{e_1}{U}} \cdot\scap{e_1}{U}}
  = \frac{\sqrt{\pi}\Gamma(\frac{m-1}{2})}{m \Gamma(\frac{m}{2})},
\end{equation}
where $U$ is uniformly drawn on the sphere $\mathbb{S}^{m-1}$, and $\Gamma$ is the gamma function.
This suggests Algorithm~\ref{sgd:alg:sgd}.

\begin{algorithm}[H]
  \caption{Median regression with SGD.}
  \KwData{A model $f_\theta$ for $\theta \in \Theta$, some data $(X_i)_{i\leq n}$, a labeling budget $T$, a step size rule $\gamma:\N\to\R_+$}
  \KwResult{A learned parameter $\hat{\theta}$ and the predictive function
    $\hat{f} = f_{\hat{\theta}}$.}

  Initialize $\theta_0$.\\
  \For{$t\gets 1$ \KwTo $T$}{
    Sample $U_t$ uniformly on $\mathbb{S}^{m-1}$.\\
    Query $\epsilon = \sign(\scap{Y_t-z}{U_t})$ for $z =
      f_{\theta_{t-1}}(X_t)$.\\
    Update the parameter
    $\theta_{t} = \theta_{t-1} + \gamma(t) \epsilon\cdot U_t^\top(Df_{\theta_{t-1}}(X_t))$.
  }
  Output $\hat{\theta} = \theta_T$, or some average, {\em e.g.}, $\hat{\theta} = T^{-1}\sum_{t=1}^T \theta_t$.
  \label{sgd:alg:sgd}
\end{algorithm}

\section{Statistical analysis}
\label{sgd:sec:stat}

In this section, we quantify the performance of Algorithm~\ref{sgd:alg:sgd} by proving optimal rates of convergence when the median regression problem is approached with (reproducing) kernels.
For simplicity, we will assume that $f^*$ can be parametrized by a linear model (potentially of infinite dimension).

\begin{assumption}
  \label{sgd:ass:source}
  Assume that the solution $f^*:\X\to\R^m$ of the median regression problem~\eqref{sgd:eq:obj} and \eqref{sgd:eq:median} can be parametrized by some separable Hilbert space ${\cal H}$, and a bounded feature map $\phi:\X\to{\cal H}$, such that, for any $i \in [m]$, there exists some $\theta_i^* \in \cal H$ such that
  \(
  \scap{f^*(\cdot)}{e_i}_{\Y} = \scap{\theta_i^*}{\phi(\cdot)}_{\cal H},
  \)
  where $(e_i)$ is the canonical basis of $\R^m$.
  Written into matrix form, there exists $\theta^* \in \Y\otimes{\cal H}$, such that
  \(
  f^*(\cdot) = \theta^* \phi(\cdot).
  \)
\end{assumption}

The curious reader can easily relax this assumption in the realm of reproducing kernel Hilbert spaces following the work of \citet{PillaudVivien2018}.
Under the linear model of Assumption~\ref{sgd:ass:source}, Algorithm~\ref{sgd:alg:sgd} is specified with
\(
u^\top Df_\theta(x) = u\otimes \phi(x).
\)
Note that rather than working with $\Theta = \Y \otimes {\cal H}$ which is potentially infinite-dimensional, empirical estimates can be represented in the finite-dimensional space $\Y \otimes \Span\brace{\phi(X_i)}_{i\leq n}$, and well approximated by small-dimensional spaces to ensure efficient computations \citep{Williams2000,Meanti2020}.

One of the key points of SGD is that gradient descent is so gradual that one can use noisy or stochastic gradients without loosing statistical guarantees while speeding up computations.
This is especially true when minimizing convex functions that are nor strongly-convex, {\em i.e.}, bounded below by a quadratic, nor smooth, {\em i.e.}, with Lipschitz-continuous gradient \citep[see, \emph{e.g.},][]{Bubeck2015}.
In particular, the following theorem, proven in Appendix~\ref{sgd:proof:sgd}, states that Algorithm~\ref{sgd:alg:sgd} minimizes the population risk at a speed at least proportional to $O(T^{-1/2})$.

\begin{theorem}[Convergence rates]
  \label{sgd:thm:sgd}
  Under Assumption~\ref{sgd:ass:source}, and under the knowledge of $\kappa$ and $M$ two real values such that $\E[\norm{\phi(X)}^2] \leq \kappa^2$ and $\norm{\theta^*} \leq M$, with a budget $T\in\N$, a constant step size $\gamma = \frac{M}{\kappa\sqrt{T}}$ and the average estimate $\hat{\theta} = \frac{1}{T}\sum_{t=0}^{T-1} \theta_t$, Algorithm~\ref{sgd:alg:sgd} leads to an estimate $f$ that suffers from an excess of risk
  \begin{equation}
    \label{sgd:eq:thm}
    \E\bracket{{\cal R}\paren{f_{\hat{\theta}}}} - {\cal R}(f^*)
    \leq \frac{2\kappa M}{c_2 \sqrt{T}}
    \leq \kappa M m^{3/2} T^{-1/2},
  \end{equation}
  where the expectation is taken with respect to the randomness of $\hat{\theta}$ that depends on the dataset $(X_i, Y_i)$ as well as the questions $(i_t, S_t)_{t\leq T}$.
\end{theorem}

While we give here a result for a fixed step size, one could retake the extensive literature on SGD to prove similar results for decaying step sizes that do not require to know the labeling budget in advance ({\em e.g.} setting $\gamma(t) \propto t^{-1/2}$ at the expense of an extra term in $\log(T)$ in front of the rates), as well as different averaging strategies \citep[see \emph{e.g.},][]{Bach2023}.
In practice, one might not know {\em a priori} the parameter $M$ but could nonetheless find the right scaling for $\gamma$ based on cross-validation.

The rate in $O(T^{-1/2})$ applies more broadly to all the strategies described in Section~\ref{sgd:sec:sgd} as long as the loss $\ell$ and the parametric model $f_\theta$ ensure that ${\cal R}(\theta)$ is convex and Lipschitz-continuous.
Although the constants appearing in front of rates depend on the complexity to reconstruct the full gradient $\nabla_\theta \ell(f_\theta(X_i, Y_i))$ from the reconstruction scheme \eqref{sgd:eq:condition}.
Those constants correspond to the second moment of the stochastic gradient.
For example, for the least-squares technique described earlier one would have to replace $c_2$ by $c_1$ in~\eqref{sgd:eq:thm}.

Theorem~\ref{sgd:thm:minmax_opt}, proven in Appendix~\ref{sgd:proof:lower}, states that any algorithm that accesses a fully supervised learning dataset of size $T$ cannot beat the rates in $O(T^{-1/2})$, hence any algorithm that collects weaker information on $(Y_i)_{i\leq T}$ cannot display better rates than the ones verified by Algorithm~\ref{sgd:alg:sgd}. This proves minimax optimality of our algorithm up to constants.

\begin{theorem}[Minimax optimality]
  \label{sgd:thm:minmax_opt}
  Under Assumption~\ref{sgd:ass:source} and the knowledge of an upper bound on $\norm{\theta^*}\leq M$, assuming that $\phi$ is bounded by $\kappa$, there exists a universal constant $c_3$ such that for any algorithm~${\cal A}$ that takes as input ${\cal D}_T = (X_i, Y_i)_{i\leq T} \sim\rho^{\otimes T}$ for any $T\in\N$ and output a parameter $\theta$,
  \begin{equation}
    \sup_{\rho\in {\cal M}_{M}} \E_{{\cal D}_T\sim\rho^{\otimes T}}\bracket{{\cal R}(f_{{\cal A}({\cal D}_T; \rho)})} - {\cal R}(f_\rho; \rho) \geq c_3 M\kappa T^{-1/2}.
  \end{equation}
  The supremum over $\rho\in{\cal M}_{M}$ has to be understood as the supremum over all distributions $\rho\in\prob{\X\times\Y}$ such that the problem defined through the risk ${\cal R}(f; \rho) := \E_{\rho}[\ell(f(X), Y)]$ is minimized for $f_\rho$ that verifies Assumption~\ref{sgd:ass:source} with $\norm{\theta^*}$ bounded by a constant $M$.
\end{theorem}

The same theorem applies for least-squares with a different universal constant.
It should be noted that minimax lower bounds are in essence quantifying worst cases of a given class of problems.
In particular, to prove Theorem~\ref{sgd:thm:minmax_opt}, we consider distributions that lead to hard problems; more specifically, we assumed the variance of the conditional distribution $\paren{Y\midvert X}$ to be high.
The practitioner should keep in mind that it is possible to add additional structure on the solution, leverage active learning or semi-supervised strategy such as uncertainty sampling \citep{Nguyen2021}, or Laplacian regularization \citep{Zhu2003,Cabannes2021c}, and reduce the optimal rates of convergence.

To conclude this section, let us remark that most of our derivations could easily be refined for practitioners facing a slightly different cost model for annotation.
In particular, they might prefer to perform batches of annotations before updating $\theta$ rather than modifying the question strategy after each input annotation.
This would be similar to mini-batching in gradient descent.
Indeed, the dependency of our result on the annotation cost model and on Assumption~\ref{sgd:ass:source} should not be seen as a limitation but rather as a proof of concept.

\section{Numerical analysis}
\label{sgd:sec:exp}

In this section, we illustrate the differences between our active method versus a classical passive method, for regression and classification problems.
Further discussions are provided in Appendix \ref{sgd:app:experiments}.
Our code is available online at \url{https://github.com/VivienCabannes/active-labeling}.

\begin{figure}[ht]
  \centering
  \includegraphics{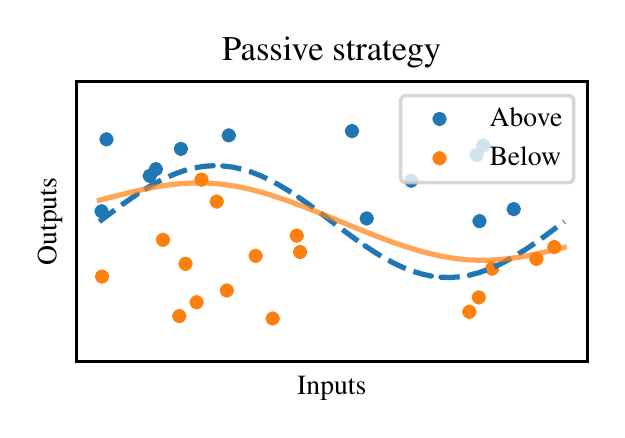}
  \includegraphics{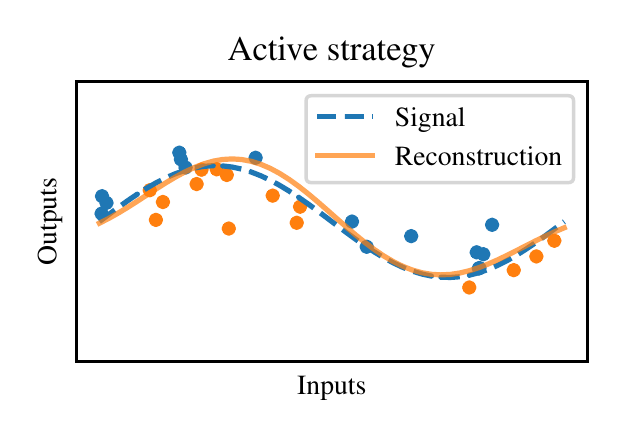}
  \caption{
    {\em Visual comparison of active and passive strategies.}
    Estimation in orange of the original signal $f^*$ in dashed blue based on median regression in a noiseless setting.
    Any orange point $(x, u)\in\R^2$ corresponds to an observation made that $u$ is below $f^*(x)$, while a blue point corresponds to $u$ above $f^*(x)$.
    The passive strategy corresponds to acquiring information based on $\paren{U\midvert x}$ following a normal distribution, while the active strategy corresponds to $\paren{u\midvert x} = f_{\theta}(x)$.
    The active strategy reconstructs the signal much better given the budget of $T=30$ observations.
  }
  \label{sgd:fig:exp_1}
\end{figure}

Let us begin with the regression problem that consists in estimating the function $f^*$ that maps $x\in[0,1]$ to $\sin(2\pi x) \in \R$.
Such a regular function, which belongs to any H\"older or Sobolev classes of functions, can be estimated with the Gaussian kernel, which would ensure Assumption \ref{sgd:ass:source}, and that corresponds to a feature map $\phi$ such that $k(x, x'):=\scap{\phi(x)}{\phi(x')} = \exp(-\abs{x-x'}/(2\sigma^2))$ for any bandwidth parameter $\sigma > 0$.\footnote{A noteworthy computational aspect of linear models, often refer as the ``kernel trick'', is that the features map $\phi$ does not need to be explicit, the knowledge of $k:\X\times\X\to\R$ being sufficient to compute all quantities of interest \citep{Scholkopf2001}. This ``trick'' can be applied to our algorithms.}
On Figure \ref{sgd:fig:exp_1}, we focus on estimating $f^*$ given data $(X_i)_{i\in [T]}$ that are uniform on $[0, 1]$ in the noiseless setting where $Y_i = f^*(X_i)$, based on the minimization of the absolute deviation loss.
The passive baseline consists in randomly choosing a threshold $U_i\sim{\cal N}(0, 1)$ and acquiring the observations $(\ind{Y_i > U_i})_{i\in [T]}$ that can be cast as the observation of the half-space $S_i = \brace{y\in\Y\midvert \ind{y > U_i} = \ind{Y_i > U_i}} =: s(Y_i, U_i)$.
In this noiseless setting, a good baseline to learn $f^*$ from the data $(X_i, S_i)$ is provided by the infimum loss characterization \citep[see][]{Cabannes2020}
\[
  f^* = \argmin_{f:\X\to\Y} \E_{(X, S)}[\inf_{y\in S} \ell(f(X), y)],
\]
where the distribution over $X$ corresponds to the marginal of $\rho$ over $\X$, and the distribution over $\paren{S\midvert X=x}$ is the pushforward of $U\sim{\cal N}(0, 1)$ under $s(f^*(x), \cdot)$.
The left plot on Figure \ref{sgd:fig:exp_1} corresponds to an instance of SGD on such an objective based on the data $(X_i, S_i)$, while the right plot corresponds to Algorithm \ref{sgd:alg:sgd}.
We take the same hyperparameters for both plots, a bandwidth $\sigma=0.2$ and an SGD step size $\gamma =0.3$.
We refer the curious reader to Figure \ref{sgd:fig:exp_1_app} in Appendix \ref{sgd:app:experiments} for plots illustrating the streaming history, and to Figure \ref{sgd:fig:exp_libsvm} for ``real-world'' experiments.

To illustrate the versatility of our method, we approach a classification problem through the median surrogate technique presented in Proposition \ref{sgd:prop:sur}.
To do so, we consider the classification problem with $m\in\N$ classes, $\X = [0,1]$ and the conditional distribution $\paren{Y\midvert X}$ linearly interpolating between Dirac in $y_1$, $y_2$ and $y_3$ respectively for $x=0$, $x=1/2$ and $x=1$ and the uniform distribution for $x=1/4$ and $x=3/4$; and $X$ uniform on $\X \setminus ([1/4 - \epsilon, 1/4+\epsilon] \cup [3/4-\epsilon, 3/4+\epsilon])$.

\begin{figure}[ht]
  \centering
  \includegraphics{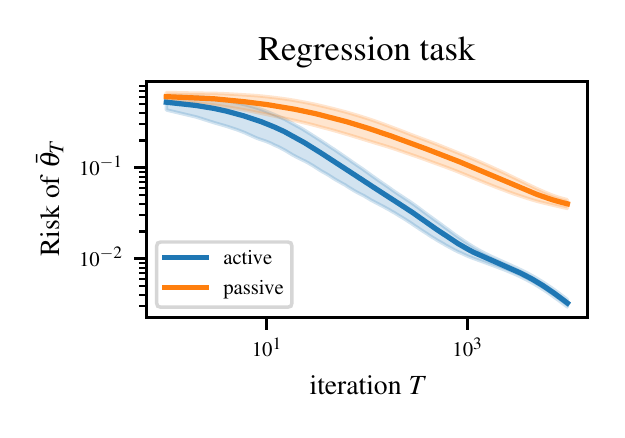}
  \includegraphics{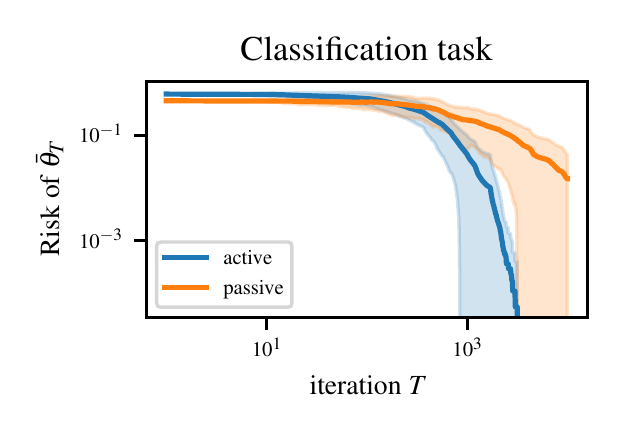}
  \caption{
    {\em Comparison of generalization errors of passive and active strategies} as a function of the annotation budget $T$.
    This error is computed by averaging over 100 trials.
    In solid is represented the average error, while the height of the dark area represents one standard deviation on each side.
    In order to consider the streaming setting where $T$ is not known in advance, we consider the decreasing step size $\gamma(t) = \gamma_0 /\sqrt{t}$; and to smooth out the stochasticity due to random gradients, we consider the average estimate $\bar\theta_t = (\theta_1 + \cdots + \theta_t) / t$.
    The left figure corresponds to the noiseless regression setting of Figure \ref{sgd:fig:exp_1}, with $\gamma_0 = 1$. We observe the convergence behavior in $O(T^{-1/2})$ of our active strategy.
    The right setting corresponds to the classification problem setting described in the main text with $m=100$, $\epsilon = 1/20$, and approached with the median surrogate.
    We observe the exponential convergence phenomenon described by \cite{PillaudVivien2018b,Cabannes2021b}; its kicks in earlier for the active strategy.
    The two plots are displayed with logarithmic scales on both axes.
  }
  \label{sgd:fig:exp_2}
\end{figure}

\section{Discussion}
\label{sgd:sec:discussion}

\subsection{Discrete output problems}
In this section, we discuss casting Algorithm \ref{sgd:alg:sgd} into a procedure to tackle discrete-output problems, by leveraging surrogate regression tasks.

Learning problems with discrete output spaces are not as well understood as regression problems.
This is a consequence of the complexity of dealing with combinatorial structures in contrast with continuous metric spaces.
In particular, gradients are not defined for discrete output models.
The current state-of-the-art framework to deal with discrete output problems is to introduce a continuous surrogate problem whose solution can be decoded as a solution on the original problem \citep{Bartlett2006}.
For example, one could solve a classification task with a median regression surrogate problem, which is the object of the next proposition, proven in Appendix~\ref{sgd:proof:sur}.

\begin{proposition}[Consistency of median surrogate]
  \label{sgd:prop:sur}
  The classification setting where $\Y$ is a finite space, and $\ell:\Y\times\Y\to\R$ is the zero-one loss $\ell(y, z) = \ind{y\neq z}$ can be solved as a regression task through the simplex embedding of $\Y$ in $\R^\Y$ with the orthonormal basis $(e_y)_{y\in\Y}$.
  More precisely, if $g^*:\X\to\R^\Y$ is the minimizer of the median surrogate risk ${\cal R}_S(g) = \E\bracket{\norm{g(X) - e_Y}}$, then $f^*:\X\to\Y$ defined as $f^*(x) = \argmax_{y\in\Y} g^*_y(x)$ minimizes the original risk ${\cal R}(f) = \E\bracket{\ell(f(X), Y)}$.\footnote{As a side note, while we are not aware of any generic theory encompassing the absolute-deviation surrogate of Proposition~\ref{sgd:prop:sur}, we showcase its superiority over least-squares on at least two types of problems on Figures \ref{sgd:fig:med_ls} and \ref{sgd:fig:med_simplex} in Appendix \ref{sgd:proof:sur}.}
\end{proposition}

More generally, any discrete output problem can be solved by reusing the consistent least-squares surrogate of \cite{Ciliberto2020}.
Algorithm \ref{sgd:alg:sgd} can be adapted to the least-squares problem based on specifications at the beginning of Section \ref{sgd:sec:median}.
This allows using our method in an off-the-shelve fashion for all discrete output problems.
For example, a problem consisting in ranking preferences over $m$ items can be approached with the Kendall correlation loss $\ell(y, z) = -\phi(y)^\top \phi(z)$ with $\phi(y) = (1_{y(i) > y(j)})$ for $i < j\leq m$, where $y$ and $z$ are permutations over $[m]$ that encode the rank of each element in terms of user preferences.
In this setting, the surrogate task introduced by \citet{Ciliberto2020} consists in learning $g(x) = \mathbb{E}[\phi(Y)\vert X=x]$ as a least-squares problem.
The half-space surrogate queries translate directly into the questions $\sum_{i < j\leq m} w(i,j) 1_{y(i) > y(j)} > c$ for some $(w(i,j)), c$ in $\mathbb{R}$. 
In particular, if $U$ is chosen to be uniform on the canonical basis (rather than on the sphere), those questions translate into pairwise orderings ({\em e.g.}, does user $x$ prefer movie $i$ or movie $j$?).
In terms of guarantee akin to Theorem \ref{sgd:thm:sgd}, retaking the calibration inequality of \citet{Ciliberto2020}, we get convergence rates of the form $m^{3/2} T^{-1/4}$.
In terms of guarantee akin to Theorem \ref{sgd:thm:minmax_opt}, since we need as least $\log_2(m!) \simeq m\log(m)$ binary queries to discriminate between $m!$ permutations, we can expect a lower bound in $m^{1/2} \log(m)^{1/2} T^{-1/2}$.
More generally, many ranking problems can be approached with correlation losses and tackled through surrogate regression problems on the convex hulls of some well-known polytopes such as the Birkhoff polytope or the permutohedron \citep[\emph{e.g.},][]{Ailon2014}.
Although their descriptions is out-of-scope of this paper, linear cuts of those polytopes form well-structured queries sets -- {\em e.g.}, the faces of all dimensions of the permutohedron correspond, in a one-to-one fashion, to strict weak orderings \citep{Ziegler1995}.

In those discrete settings, Theorem \ref{sgd:thm:sgd} can be refined under low noise conditions.
In particular, under generalization of the Massart noise condition, our approach could even exhibit exponential convergence rates as illustrated on Figure \ref{sgd:fig:exp_2}.
For classification problems, this condition can be expressed as the existence of a threshold $\delta > 0$ such that for almost all $x\in{\cal X}$ and $z\in {\cal Y}$, we have $\mathbb{P}(Y = f(x) \vert X=x) - \mathbb{P}(Y = z\vert X=x) \notin (0, \delta)$.
Arguably, this assumption is met on well-curated images dataset such as ImageNet or CIFAR10, where for each input $X$ the most probable class has always more than {\em e.g.} 60\% of chance to be the target $Y$.
When this assumption holds together with Assumption \ref{sgd:ass:source} (when the surrogate target $g^*$ belongs to the RKHS and the kernel is bounded), then the right hand-side of equation \eqref{sgd:eq:thm} can be replaced by $\exp(-cT)$ for some constant $c$.
The proof would be a simple adaptation of \citet{PillaudVivien2018,Cabannes2021b} to our case.

\subsection{Supervised learning baseline with resampling}
In this section, we discuss simple supervised learning baselines that compete with Algorithm \ref{sgd:alg:sgd} when resampling is allowed.

When resampling is allowed a simple baseline for the active labeling problem is provided by supervised learning.
In regression problems with the query of any half-space, a method that consists in annotating each $(Y_i)_{i\leq n(T, \epsilon)}$ up to precision $\epsilon$, before using any supervised learning method to learn $f$ from $(X_i, Y_i)_{i\leq n(T, \epsilon)}$ could acquire $n(T, \epsilon) \simeq T / m \log_2(\epsilon^{-1})$ data points with a dichotomic search along all directions, assuming $Y_i$ bounded or sub-Gaussian.
In terms of minimax rates, such a procedure cannot perform better than in $n(T,\epsilon)^{-1/2} + \epsilon$, the first term being due to the statistical limit in Theorem~\ref{sgd:thm:minmax_opt}, the second due to the incertitude $\epsilon$ on each $Y_i$ that transfers to the same level of incertitude on $f$.
Optimizing with respect to $\epsilon$ yields a bound in $O(T^{-{1/2}}\log(T)^{1/2})$.
Therefore, this not-so-naive baseline is only suboptimal by a factor $\log(T)^{1/2}$.
In the meanwhile, Algorithm \ref{sgd:alg:sgd} can be rewritten with resampling, as well as Theorem \ref{sgd:thm:sgd}, which we prove in Appendix \ref{sgd:proof:resampling}.
Hence, our technique will still achieve minimax optimality for the problem ``with resampling''.
In other terms, by deciding to acquire more imprecise information, our algorithm reduces annotation cost for a given level of generalization error (or equivalently reduces generalization error for a given annotation budget) by a factor $\log(T)^{1/2}$ when compared to this baseline.

The picture is slightly different for discrete-output problems.
If one can ask any question $s\in 2^\Y$ then with a dichotomic search, one can retrieve any label with $\log_2(m)$ questions.
Hence, to theoretically beat the fully supervised baseline with the SGD method described in Section~\ref{sgd:sec:sgd}, one would have to derive a gradient strategy \eqref{sgd:eq:sgd_def} with a small enough second moment ({\em e.g.}, for convex losses that are non-smooth nor strongly convex, the increase in the second moment compared to the usual stochastic gradients should be no greater than $\log_2(m)^{1/2}$).
How to best refine our technique to better take into account the discrete structure of the output space is an open question.
Introducing bias that does not modify convergence properties while reducing variance eventually thanks to importance sampling is a potential way to approach this problem.
A simpler idea would be to remember information of the type $Y_i \in s$ to restrict the questions asked in order to locate $f_{\theta_t}(X_i) - Y_i$ when performing stochastic gradient descent with resampling.
Combinatorial bandits might also provide helpful insights on the matter.
Ultimately, we would like to build an understanding of the whole distribution $\paren{Y\midvert X}$ and not only of $f^*(X)$ as we explore labels in order to refine this exploration.

\subsection{Min-max approaches}
In this section, we discuss potential extensions of our SGD procedure, based on min-max variational objectives.

Min-max approaches have been popularized for searching games and active learning, where one searches for the question that minimizes the size of the space where a potential guess could lie under the worst possible answer to that question.
A particularly well illustrative example is the solution of the Mastermind game proposed by \cite{Knuth1977}.
While our work leverages plain SGD, one could build on the vector field point-of-view of gradient descent \citep[see, \emph{e.g.},][]{Bubeck2015} to tackle min-max convex concave problems with similar guarantees.
In particular, we could design weakly supervised losses $L(f(x), s; \ind{y\in s})$ and min-max games where a prediction player aims at minimizing such a loss with respect to the prediction $f$, while the query player aims at maximizing it with respect to the question $s$, that is querying information that best elicit mistakes made by the prediction player.
For example, the dual norm characterization of the norm leads to the following min-max approach to the median regression
\[
  \argmin_{f:\X\to\Y}{\cal R}(f) = \argmin_{f:\X\to\Y}\max_{U\in(\mathbb{S}^{m-1})^{\X\times\Y}} \E_{(X, Y)\sim\rho}\bracket{\scap{U(x, y)}{f(x) - y}}.
\]
Such min-max formulations would be of interest if they lead to improvement of computational and statistical efficiencies, similarly to the work of \cite{Babichev2019}.
For classification problems, the following proposition introduces such a game and suggests its suitability.
Its proof can be found in Appendix \ref{sgd:proof:minmax}.

\begin{proposition}
  \label{sgd:prop:minmax}
  Consider the classification problem of learning $f^*:\X\to\Y$ where $\Y$ is of finite cardinality, with the 0-1 loss $\ell(z, y) = \ind{z\neq y}$, minimizing the risk \eqref{sgd:eq:obj} under a distribution $\rho$ on $\X\times\Y$.
  Introduce the surrogate score functions $g:\X\to\prob{\Y}; x\to v$ where $v = (v_y)_{y\in \Y}$ is a family of non-negative weights that sum to one, as well as the surrogate loss function $L:\prob{\Y}\times{\cal S}\times\brace{-1, 1}\to \R; (v, S, \epsilon) = \epsilon(1 - 2\sum_{y\in S} v_y)$, and the min-max game
  \begin{equation}
    \label{sgd:eq:minmax}
    \min_{g:\X\to\prob{\Y}} \max_{\mu:\X\to\prob{\cal S}} \E_{(X,Y)\sim\rho} \E_{S\sim \mu(x)}\bracket{L(g(x), S; \ind{Y\in S}-\ind{Y\notin S})}.
  \end{equation}
  When ${\cal S}$ contains the singletons and with the low-noise condition that $\Pbb\paren{Y\neq f^*(x)\midvert X=x} < 1/2$ almost everywhere, then $f^*$ can be learned through the relation $f^*(x) = \argmin_{y\in\Y} g^*(x)_y$ for the unique minimizer $g^*$ of \eqref{sgd:eq:minmax}.
  Moreover, the minimization of the empirical version of this objective with the stochastic gradient updates for saddle point problems provides a natural ``active labeling'' scheme to find this $g^*$.
\end{proposition}

On the one hand, this min-max formulation could help to easily incorporate restrictions on the sets to query.
On the other hand, it is not completely clear how to best update (or derive an unbiased stochastic gradient strategy for) the adversarial query strategy $\mu$ based on partial information.

\section{Conclusion}
We have introduced the ``active labeling'' problem, which corresponds to ``active partially supervised learning''.
We provided a solution to this problem based on stochastic gradient descent.
Although our method can be used for any discrete output problem, we detailed how it works for median regression, where we show that it optimizes the generalization error for a given annotation budget.
In a near future, we would like to focus on better exploiting the discrete structure of classification problems, eventually with resampling strategies.

Understanding more precisely the key issues in applications concerned with privacy, and studying how weak gradients might provide a good trade-off between learning efficiently and revealing too much information also provide interesting follow-ups.
Finally, regarding dataset annotation, exploring different paradigms of weakly supervised learning would lead to different active weakly supervised learning frameworks.
While this work is based on partial labeling, similar formalization could be made based on other weak supervision models, such as aggregation \citep[\emph{e.g.},][]{Ratner2020}, or group statistics \citep{Dietterich1997}.
In particular, annotating a huge dataset is often done by bagging inputs according to predicted labels and correcting errors that can be spotted on those bags of inputs \citep{ImageNet}.
We left for future work the study of variants of the ``active labeling'' problem that model those settings.

\begin{ack}
  While at INRIA / ENS / PSL, VC was funded in part by the French government under management of Agence Nationale de la Recherche as part of the ``Investissements d'avenir'' program, reference ANR-19-P3IA-0001 (PRAIRIE 3IA Institute). 
  FR and AR also acknowledges support of the European Research Council (grants SEQUOIA 724063 and REAL 947908).
\end{ack}

\bibliography{main}

\section*{Checklist}
\begin{enumerate}

\item For all authors...
\begin{enumerate}
  \item Do the main claims made in the abstract and introduction accurately reflect the paper's contributions and scope?
    \answerYes{}
  \item Did you describe the limitations of your work?
    \answerYes{See discussion section.}
  \item Did you discuss any potential negative societal impacts of your work?
    \answerNA{This work aims at developping advanced techniques to learn without too much supervision. Such a quest of increasing AI systems capability at a reduced human labor cost is associated with broad societal issues. Those questions being really generic, we did not mention them in the main text.}
  \item Have you read the ethics review guidelines and ensured that your paper conforms to them?
    \answerYes{}
\end{enumerate}

\item If you are including theoretical results...
\begin{enumerate}
  \item Did you state the full set of assumptions of all theoretical results?
    \answerYes{}
  \item Did you include complete proofs of all theoretical results?
    \answerYes{}
\end{enumerate}

\item If you ran experiments...
\begin{enumerate}
  \item Did you include the code, data, and instructions needed to reproduce the main experimental results (either in the supplemental material or as a URL)?
    \answerYes{}
  \item Did you specify all the training details (e.g., data splits, hyperparameters, how they were chosen)?
    \answerYes{}
        \item Did you report error bars (e.g., with respect to the random seed after running experiments multiple times)?
    \answerYes{}
        \item Did you include the total amount of compute and the type of resources used (e.g., type of GPUs, internal cluster, or cloud provider)?
    \answerNA{The experiments were run on a personal laptop and did not require many charges. Indeed, the amount of compute for experiments were similar to the amount used to write this paper.}
\end{enumerate}

\item If you are using existing assets (e.g., code, data, models) or curating/releasing new assets...
\begin{enumerate}
  \item If your work uses existing assets, did you cite the creators?
    \answerYes{Although we have not cited the creators of some \LaTeX\ libraries we used such as Michael Sharpe and the {\tt newtx} package which we used for fonts in our text.}
  \item Did you mention the license of the assets?
    \answerNA{{\em Numpy} and {\em LIBSVM} are under Berkeley Software Distribution licenses (respectively the liberal and revised ones), {\em Python} and {\em matplotlib} are under the Python Software Foundation license.}
  \item Did you include any new assets either in the supplemental material or as a URL?
    \answerYes{}
  \item Did you discuss whether and how consent was obtained from people whose data you're using/curating?
    \answerNA{}
  \item Did you discuss whether the data you are using/curating contains personally identifiable information or offensive content?
    \answerNA{}
\end{enumerate}

\item If you used crowdsourcing or conducted research with human subjects...
\begin{enumerate}
  \item Did you include the full text of instructions given to participants and screenshots, if applicable?
    \answerNA{}
  \item Did you describe any potential participant risks, with links to Institutional Review Board (IRB) approvals, if applicable?
    \answerNA{}
  \item Did you include the estimated hourly wage paid to participants and the total amount spent on participant compensation?
    \answerNA{}
\end{enumerate}

\end{enumerate}
\clearpage

\appendix

\section{Proofs of the statistical analysis}
In the following proofs, we assume $\X$ to be Polish and $\Y = \R^m$, so to define the joint probability $\rho\in\prob{\X\times\Y}$.
Moreover, we assume that $\E[\norm{Y}] < +\infty$ in order to define the risk of median regression.
We consider ${\cal H}$ to be a Hilbert space that is separable ({\em i.e.} only the origin is in all the neighborhood of the origin), and $\phi$ to be a measurable mapping from $\X$ to ${\cal H}$.

In terms of notations, we denote $\brace{1, 2, \cdots, n}$ by $[n]$ for any $n\in\N^*$, and by $(x_i)_{i\leq n}$ the family $(x_1, \cdots, x_n)$ for any sequence $(x_i)$.
The unit sphere in $\R^m$ is denoted by $\mathbb{S}^{m-1}$.
The symbol $\otimes$ denotes tensors, and is extended to product measures in the notation $\rho^{\otimes n} = \rho\times\rho\times\cdots\times\rho$.
We have used the isometry between trace-class linear mappings from ${\cal H}$ to $\Y$ and the tensor space $\Y\otimes{\cal H}$, which generalizes the matrix representation of linear map between two finite-dimensional vector spaces.
This space inherits from the Hilbertian structure of ${\cal H}$ and $\Y$ and we denote by $\norm{\cdot}$ the Hilbertian norm that generalizes the Frobenius norm on linear maps between Euclidean spaces.

\subsection{Upper bound for stochastic gradient descent}
\label{sgd:proof:sgd}

This subsection is devoted to the proof of Theorem \ref{sgd:thm:sgd}.
For simplicity, we will work with the rescaled step size $\gamma_t := c_2 \gamma(t)$ rather than the step size described in the main text $\gamma(t)$.

Convergence of stochastic gradient descent for non-smooth problems is a known result. For completeness, we reproduce and adapt a usual proof to our setting.
For $t\in\N$, let us introduce the random functions
\[
	{\cal R}_t(\theta) = c_2^{-1}\abs{\scap{\theta\phi(X_t) - Y_t}{U_t}},
	\qquad\text{where}\qquad
	c_2 = \E_U[\abs{\scap{e_1}{U}}] = \E_U[\sign(\scap{e_1}{U}) \scap{e_1}{U}]
\]
for $(X_t, Y_t) \sim \rho$, $U_t$ uniform on the sphere $\mathbb{S}^{m-1} \subset \Y$.
Those random functions all average to ${\cal R}(\theta) = \E_{\rho} \E_U[c_2^{-1}\abs{\scap{\theta\phi(X) - Y}{U}}] = \E_\rho[\norm{\theta\phi(X) - Y}]$.
After a random initialization $\theta_0 \in\Theta$, the stochastic gradient update rule can be written for any $t \in \N$ as
\[
	\theta_{t+1} = \theta_t - \gamma_t \nabla {\cal R}_t(\theta_t),
\]
where $\nabla {\cal R}_t$ denotes any sub-gradients of ${\cal R}_t$.
We can compute
\[
	\nabla {\cal R}_t(\theta_t)
	= c_2^{-1} \nabla \abs{\scap{\theta\phi(X_t) - Y_t}{U_t}}
	= c_2^{-1} \sign\paren{\scap{\theta\phi(X_t) - Y_t}{U_t}} U_t\otimes \phi(X_t).
\]
This corresponds to the gradient written in Algorithm \ref{sgd:alg:sgd}.

Let us now express the recurrence relation on $\norm{\theta_{t+1} - \theta^*}$.
We have
\begin{align*}
	\norm{\theta_{t+1} - \theta^*}^2
	 & = \norm{\theta_t - \gamma_t \nabla {\cal R}_t(\theta_t) - \theta^*}^2
	\\&= \norm{\theta_t - \theta^*}^2 + \gamma_t^2\norm{\nabla {\cal R}_t(\theta_t)}^2 - 2\gamma_t\scap{\nabla {\cal R}_t(\theta_t)}{\theta_t - \theta^*}.
\end{align*}
Because ${\cal R}_t$ is convex, it is above its tangents
\[
	{\cal R}_t(\theta^*) \geq {\cal R}_t(\theta_t) + \scap{\nabla{\cal R}_t(\theta_t)}{\theta^* - \theta_t}.
\]
Hence,
\[
	\norm{\theta_{t+1} - \theta^*}^2
	\leq \norm{\theta_t - \theta^*}^2 + \gamma_t^2\norm{\nabla {\cal R}_t(\theta_t)}^2 + 2\gamma_t ({\cal R}_t(\theta^*) - {\cal R}_t(\theta_t)).
\]
This allows bounding the excess of risk as
\[
	2({\cal R}_t(\theta_t) - {\cal R}_t(\theta^*))
	\leq \frac{1}{\gamma_t}(\norm{\theta_t - \theta^*}^2 - \norm{\theta_{t+1} - \theta^*}^2) + \gamma_t c_2^{-2}\norm{\phi(X_t)}^2.
\]
where we used the fact that $\norm{\nabla {\cal R}_t} = c_2^{-1}\norm{\phi(X_t)}$.
Let us multiply this inequality by $\eta_t > 0$ and sum from $t=0$ to $t=T-1$, we get
\begin{align*}
	 & 2(\sum_{t=0}^{T-1} \eta_t {\cal R}_t(\theta_t) - \sum_{t=0}^{T-1} \eta_t {\cal R}_t(\theta^*))
	\leq \sum_{t=0}^{T-1} \frac{\eta_t}{\gamma_t}(\norm{\theta_t - \theta^*}^2 - \norm{\theta_{t+1} - \theta^*}^2) + \sum_{t=0}^{T-1} \eta_t\gamma_t c_2^{-2}\norm{\phi(X_t)}^2
	\\&\qquad= \frac{\eta_0}{\gamma_0} \norm{\theta_0-\theta^*}^2 - \frac{\eta_{T-1}}{\gamma_{T-1}} \norm{\theta_T - \theta^*}^2 + \sum_{t=1}^{T-1} \paren{\frac{\eta_t}{\gamma_t} - \frac{\eta_{t-1}}{\gamma_{t-1}}}\norm{\theta_t - \theta^*}^2 + \sum_{t=0}^{T-1} \eta_t\gamma_t c_2^{-2}\norm{\phi(X_t)}^2.
\end{align*}
From here, there is several options to obtain a convergence result, either one assume $\norm{\theta_t - \theta^*}$ bounded and take $\eta_t\gamma_{t-1} \geq \eta_{t-1}\gamma_t$; or one take $\eta_t = \gamma_t$ but at the price of paying an extra $\log(T)$ factor in the bound; or one take $\gamma_t$ and $\eta_t$ independent of $t$.
Since we suppose the annotation budget given, we will choose $\gamma_t$ and $\eta_t$ independent of $t$, only depending on $T$.
\begin{align*}
	 & 2(\sum_{t=0}^{T-1} \eta {\cal R}_t(\theta_t) - \sum_{t=0}^{T-1} \eta {\cal R}_t(\theta^*))
	\leq \frac{\eta}{\gamma} \norm{\theta_0-\theta^*}^2 + \sum_{t=0}^{T-1} \eta\gamma c_2^{-2}\norm{\phi(X_t)}^2.
\end{align*}

Let now take the expectation with respect to all the random variables, for the risk
\begin{align*}
	\E_{(X_s, Y_s, U_s)_{s\leq t}}[{\cal R}_t(\theta_t)]
	 & = \E_{(X_s, Y_s, U_s)_{s\leq t}}\bracket{\E_{(X_t, Y_t)}\bracket{\E_{U_t}\bracket{{\cal R}_t(\theta_t)\midvert \theta_t}\midvert \theta_t}}
	\\&= \E_{(X_s, Y_s, U_s)_{s\leq t}}\bracket{{\cal R}(\theta_t)}
	= \E[{\cal R}(\theta_t)].
\end{align*}
For the variance, $\E[\norm{\phi(X_s)}^2] = \E[\norm{\phi(X)}^2] = \kappa^2$.

Let us fix $T$ and consider $\eta_t = 1/ T$, by Jensen we can bound the following averaging
\begin{align*}
	2\paren{{\cal R}\paren{\sum_{t=0}^{T-1} \eta_t \theta_t} - {\cal R}(\theta^*)}
	 & \leq 2\paren{\sum_{t=0}^{T-1} \eta_t {\cal R}\paren{\theta_t} - {\cal R}(\theta^*)}
	= 2\E\bracket{\sum_{t=0}^{T-1} \eta_t ({\cal R}_t\paren{\theta_t} - {\cal R}_t(\theta^*))}
	\\&\qquad\leq \frac{1}{T\gamma}\norm{\theta_0-\theta^*}^2 + \gamma c_2^{-2}\kappa^2.
\end{align*}
Initializing $\theta_0$ to zero, we can optimize the resulting quantity to get the desired result.

\subsection{Upper bound for resampling strategy}
\label{sgd:proof:resampling}

For resampling strategies, the proof is built on classical statistical learning theory considerations.
Let us decompose the risk between estimation and optimization errors.
Recall the expression of the risk ${\cal R}$, the function taking as inputs measurable functions from $\X$ to $\Y$ and outputting a real number
\[
	{\cal R}(f) = \E_{\rho}[\norm{f(X) - Y}].
\]
Let us denote by ${\cal F}$ the class of functions from $\X$ to $\Y$ we are going to work with.
Let $f_n$ be our estimate of $f^*$ which maps almost every $x\in\X$ to the geometric median of $\paren{Y\midvert X}$.
Denote by ${\cal R}_{{\cal D}_n}^*$ the best value that can be achieved by our class of functions to minimize the empirical average absolute deviation
\[
	{\cal R}^*_{{\cal D}_n} = \inf_{f\in{\cal F}} {\cal R}_{{\cal D}_n}(f).
\]
Assumption \ref{sgd:ass:source} states that we have a well-specified model ${\cal F}$ to estimate the median, {\em i.e.} $f^* \in {\cal F}$.
Hence, the excess of risk can be decomposed as an estimation and an optimization error, without approximation error (it is not difficult to add an approximation error, but it will make the derivations longer and the convergence rates harder to parse for the reader).
Using the fact that ${\cal R}_{{\cal D}_n}(f^*) \geq {\cal R}_{{\cal D}_n}^*$ by definition of the infimum, we have
\begin{equation}
	\label{sgd:eq:decomposition}
	{\cal R}(f_n) - {\cal R}(f^*)
	\leq \underbrace{{\cal R}(f_n) - {\cal R}_{{\cal D}_n}(f_n)
	+ {\cal R}_{{\cal D}_n}(f^*) - {\cal R}(f^*)}_{\text{estimation error}}
	+ \underbrace{{\cal R}_{{\cal D}_n}(f_n) - {\cal R}_{{\cal D}_n}^*}_{\text{optimization error}}.
\end{equation}

\paragraph{Estimation error.}
Let us begin by controlling the estimation error.
We have two terms in it.
${\cal R}_{{\cal D}_n}(f^*) - {\cal R}(f^*)$ can be controlled with a concentration inequality on the empirical average of $\norm{f^*(X) - Y}$ around its population mean.
Assuming sub-Gaussian moments of $Y$, it can be done with Bernstein inequality.

${\cal R}_{{\cal D}_n}(f_n) - {\cal R}(f_n)$ is harder to control as $f_n$ depends on ${\cal D}_n$, so we can not use the same technique.
The classical technique consists in going for the brutal uniform majoration,
\begin{equation}
	\label{sgd:eq:sup_rade}
	{\cal R} (f_n) - {\cal R}_{{\cal D}_n}(f_n) \leq
	\sup_{f\in{\cal F}} \paren{{\cal R} (f) - {\cal R}_{{\cal D}_n}(f)},
\end{equation}
where ${\cal F}$ denotes the set of functions that $f_n$ could be in concordance with our algorithm.
While this bound could seem highly suboptimal, when the class of functions is well-behaved, we can indeed control the deviation ${\cal R}(f) - {\cal R}_{{\cal D}_n}(f)$ uniformly over this class without losing much (indeed for any class of functions, it is possible to build some really adversarial distribution $\rho$ so that this supremum behaves similarly to the concentration we are looking for \citep{Vapnik1995,Anthony1999}).
This is particularly the case for our model linked with Assumption \ref{sgd:ass:source}.
Expectations of supremum processes have been extensively studied, allowing to get satisfying upper bounds (note that when the $\norm{f(X) - Y}$ is bounded, deviation of the quantity of interest around its expectation can be controlled through McDiarmid inequality).
In the statistical learning literature, it is usual to proceed with Rademacher complexity.

\begin{lemma}[Uniform control of functions deviation with Rademacher complexity]
	The expectation of the excess of risk can be bounded as
	\begin{equation}
		\label{sgd:eq:rademacher}
		\frac{1}{2}\E_{{\cal D}_n} \bracket{\sup_{f\in{\cal F}} \paren{{\cal R} (f) - {\cal R}_{{\cal D}_n}(f)}}
		\leq \mathfrak{R}_n({\cal F}, \ell, \rho) :=
		\frac{1}{n} \E_{{\cal D}_n, (\sigma_i)}\bracket{ \sup_{f\in{\cal F}}\sigma_i \ell(f(X_i), Y_i)},
	\end{equation}
	where $(\sigma_i)_{i\leq n}$ is defined as a family of Bernoulli independent variables taking value one or minus one with equal probability, and $\mathfrak{R}_n({\cal F}, \ell, \rho)$ is called Rademacher complexity.
\end{lemma}

\begin{proof}
	This results from the reduction to larger supremum and a symmetrization trick,
	\begin{align*}
		\E_{{\cal D}_n} \bracket{\sup_{f\in{\cal F}} \paren{{\cal R} (f) - {\cal R}_{{\cal D}_n}(f)}}
		 & = \E_{{\cal D}_n} \bracket{\sup_{f\in{\cal F}} \paren{\E_{{\cal D}_n'}{\cal R}_{{\cal D}_n'} (f) - {\cal R}_{{\cal D}_n}(f)}}
		\\&\leq \E_{{\cal D}_n} \E_{{\cal D}_n'}\bracket{\sup_{f\in{\cal F}} \paren{{\cal R}_{{\cal D}_n'} (f) - {\cal R}_{{\cal D}_n}(f)}}
		\\&= \E_{(X_i, Y_i), (X_i', Y_i')}\bracket{\sup_{f\in{\cal F}} \paren{\frac{1}{n} \sum_{i=1}^n \ell(f(X_i'), Y_i') - \ell(f(X_i), Y_i)}}
		\\&= \E_{(X_i, Y_i), (X_i', Y_i'), (\sigma_i)}\bracket{\sup_{f\in{\cal F}} \paren{\frac{1}{n} \sum_{i=1}^n \sigma_i\paren{\ell(f(X_i'), Y_i') - \ell(f(X_i), Y_i)}}}
		\\&\leq 2 \E_{(X_i, Y_i), (\sigma_i)}\bracket{\sup_{f\in{\cal F}} \paren{\frac{1}{n} \sum_{i=1}^n \sigma_i\paren{\ell(f(X_i), Y_i)}}},
	\end{align*}
	which ends the proof.
\end{proof}

In our case, we want to compute the Rademacher complexity for $\ell$ given by the norm of $\Y$, and ${\cal F} = \brace{x\to \theta \phi(x)\midvert \theta \in \Y\otimes{\cal H}, \norm{\theta} < M}$, for $M > 0$ a parameter to specify in order to make sure that $\norm{\theta^*} < M$, where the norm has to be understood as the $\ell^2$-product norm on $\Y \otimes{\cal H} \simeq {\cal H}^m$.
Working with linear models and Lipschitz losses is a well-known setting, allowing to derive directly the following bound.

\begin{lemma}[Rademacher complexity of linear models with Lipschitz losses]
	The complexity of the linear class of vector-valued function ${\cal F} = \brace{x\to \theta \phi(x)\midvert \theta \in \Y\otimes{\cal H}, \norm{\theta} < M}$ is bounded as
	\begin{equation}
		\E_{(\sigma_i)}\bracket{\sup_{f\in{\cal F}}\paren{\frac{1}{n} \sum_{i=1}^n \sigma_i \norm{f(x_i) - y_i}}}
		\leq M \kappa n^{-1/2}.
	\end{equation}
\end{lemma}
\begin{proof}
	This proposition is usually split in two.
	First using the fact that the composition of a space of functions with a Lipschitz function does not increase the entropy of the subsequent space \citep{Vitushkin1954}.
	Then bounding the Rademacher complexity of linear models.
	We refer to \cite{Maurer2016} for a self-contained proof of this result (stated in its Section 4.3).
\end{proof}

Adding all the pieces together we have proven the following proposition, using the fact that the previous bound also applies to $\sup_{f\in{\cal F}} {\cal R}_{{\cal D}_n}(f) - {\cal R}(f)$ by symmetry, hence it can be used for the deviation of ${\cal R}_{{\cal D}_n}(f^*) - {\cal R}(f^*)$.

\begin{proposition}[Control of the estimation error]
	Under Assumption \ref{sgd:ass:source}, with the model of computation ${\cal F} = \brace{x\in\X\to \theta\phi(x) \in \Y \midvert \norm{\theta} \leq M}$, the generalization error of $f_n$ is controlled by a term in $n^{-1/2}$ plus an optimization error on the empirical risk minimization
	\begin{equation}
		\E_{{\cal D}_n}\bracket{{\cal R}(f_n) - {\cal R}(f^*)}
		\leq \frac{4 M\kappa}{n^{1/2}} + \E_{{\cal D}_n}\bracket{{\cal R}_{{\cal D}_n}(f_n) - {\cal R}_{{\cal D}_n}^*},
	\end{equation}
	as long as $f^* \in {\cal F}$.
\end{proposition}

Note that this result can be refined using regularized risk \citep{Sridharan2008}, which would be useful under richer (stronger or weaker) source assumptions \citep[\emph{e.g.},][]{Caponnetto2007}.
Such a refinement would allow switching from a constraint $\norm{\theta} < M$ to define ${\cal F}$ to a regularization parameter $\lambda\norm{\theta}^2$ added in the risk without restrictions on $\norm{\theta}$, which would be better aligned with the current practice of machine learning.
Under Assumption \ref{sgd:ass:source}, this will not fundamentally change the result.
The estimation error can be controlled with the derivation in Appendix \ref{sgd:proof:sgd}, where stochastic gradients correspond to random sampling of a coefficient $i_t\leq n$ plus the choice of a random $U_t$.
For the option without resampling, there exists an acceleration scheme specific to different losses in order to benefit from the strong convexity \citep[\emph{e.g.},][]{Bach2013}.

\subsection{Lower bound}
\label{sgd:proof:lower}

In this section, we prove Theorem \ref{sgd:thm:minmax_opt}.
Let us consider any algorithm ${\cal A}:\cup_{n\in\N}(\X\times\Y)^n \to \Theta$ that matches a dataset ${\cal D}_n$ to an estimate $\theta_{{\cal D}_n}\in\Theta$.
Let us consider jointly a distribution $\rho$ and a parameter $\theta$ such that Assumption \ref{sgd:ass:source} holds, that is $f_\rho := \argmin_{f:\X\to\Y} \E_\rho[\ell(f(X), Y)] = f_\theta$.
We are interested in characterizing for each algorithm the worst excess of risk it can achieve with respect to an adversarial distribution.
The best worst performance that can be achieved by algorithms matching datasets to parameter can be written as
\begin{equation}
	{\cal E} = \inf_{\cal A}\sup_{\theta\in\Theta, \rho\in\prob{\X\times\Y}; f_{\rho} = f_{\theta}} \E_{{\cal D}_n \sim \rho^{\otimes n}}\bracket{\E_{(X, Y)\sim\rho}\bracket{\ell(f_{{\cal A}({\cal D}_n)}(X), Y) - \ell(f_{\theta}(X), Y)}}.
\end{equation}
This provides a lower bound to upper bounds such as~\eqref{sgd:eq:thm} that can be derived for any algorithm.
There are many ways to get lower bounds on this quantity.
Ultimately, we want to quantify the best certainty one can have on an estimate $\theta$ based on some observations $(X_i, Y_i)_{i\leq n}$.
In particular, the algorithms ${\cal A}$ can be seen as rules to discriminate a model $\theta$ from observations ${\cal D}_n$ made under $\rho_\theta$, and where the error is measured through the excess of risk ${\cal R}(f_{\hat\theta}, \rho_\theta) - {\cal R}(f_\theta; \rho_\theta)$ where ${\cal R}(f; \rho) = \E_\rho[\ell(f(X), Y)]$ and $\rho_\theta$ is a distribution parametrized by $\theta$ such that $f_\theta = f_\rho$.

Let us first characterize the measure of error.
Surprisingly, when in presence of Gaussian noise or uniform noise, the excess of risk behaves like a quadratic metric between parameters.

\begin{lemma}[Quadratic behavior of the median regression excess of risk with Gaussian noise]
	Consider the random variable $Y \sim {\cal N}(\mu, \sigma^2 I_m)$, denote by $\hat{\mu}$ an estimate of $\mu$, the excess of risk can be developed as
	\begin{equation}
		\E_{{\cal N}(\mu, \sigma^2 I_m)}[\norm{\hat{\mu} - Y} - \norm{\mu - Y}]
		= \frac{c_4\norm{\hat{\mu} - \mu}^2}{\sigma}
		+ o\paren{\frac{\norm{\hat{\mu} - \mu}^3}{\sigma^2}},
	\end{equation}
	where $c_4 = \Gamma(\frac{m+1}{2}) / (2\sqrt{2}\Gamma(\frac{m+2}{2})) \geq (m+2)^{-1/2} / 2$.
\end{lemma}
\begin{proof}
	With this specific noise model, one can do the following derivations.
	\[
		\E_{{\cal N}(\mu, \sigma^2 I_m)}[\norm{\hat{\mu} - Y}]
		= \E_{{\cal N}(0, I_m)} [\norm{\hat{\mu} - \mu - \sigma Y}]
		= \sigma \E_{{\cal N}(0, I_m)} \bracket{\norm{\frac{\hat{\mu} - \mu}{\sigma} - Y}}.
	\]
	We recognize the mean of a non-central $\chi$-distribution of parameter $k=m$ and $\lambda = \norm{\frac{\hat{\mu} - \mu}{\sigma}}$.
	It can be expressed through the generalized Laguerre functions, which allows us to get the following Taylor expansion
	\begin{align*}
		\E_{{\cal N}(\mu, \sigma^2 I_m)}[\norm{\hat{\mu} - Y}]
		 & = \frac{\sqrt{\pi}\sigma}{\sqrt{2}} L_{\frac{1}{2}}^{(\frac{m-2}{2})}\paren{-\frac{\norm{\hat{\mu} - \mu}^2}{2\sigma^2}}
		\\&= \frac{\sqrt{\pi} \sigma}{\sqrt{2}} \paren{L_{\frac{1}{2}}^{(\frac{m-2}{2})}(0) + \frac{\norm{\hat{\mu} - \mu}^2}{2\sigma^2} L_{-\frac{1}{2}}^{(\frac{m}{2})}(0)}
		+ o\paren{\frac{\norm{\hat{\mu} - \mu}^3}{\sigma^2}}.
	\end{align*}
	Hence, the following expression of the excess of risk,
	\begin{align*}
		\E_{{\cal N}(\mu, \sigma^2 I_m)}[\norm{\hat{\mu} - Y} - \norm{\mu - Y}]
		 & = \frac{\sqrt{\pi}\norm{\hat{\mu} - \mu}^2}{2\sqrt{2}\sigma} L_{-\frac{1}{2}}^{(\frac{m}{2})}(0)
		+ o\paren{\frac{\norm{\hat{\mu} - \mu}^3}{\sigma^2}}
		\\&= \frac{\Gamma(\frac{m+1}{2})\norm{\hat{\mu} - \mu}^2}{2\sqrt{2}\Gamma(\frac{m+2}{2})\sigma}
		+ o\paren{\frac{\norm{\hat{\mu} - \mu}^3}{\sigma^2}}.
	\end{align*}
	Note that in dimension one, the calculation can be done explicitly by computing integrals with the error function.
	\begin{align*}
		\E_{{\cal N}(\mu, \sigma^2)}[\norm{\hat{\mu} - Y}]
		 & = \sigma \E_{{\cal N}(0, 1)} \bracket{Y - \frac{\hat{\mu} - \mu}{\sigma} + 2\ind{Y < \frac{\hat{\mu} - \mu}{\sigma}} \paren{\frac{\hat{\mu} - \mu}{\sigma} - Y}}
		\\&= \mu - \hat{\mu} + 2(\hat{\mu} - \mu) \E_{{\cal N}(0, 1)}\bracket{\ind{Y < \frac{\hat{\mu} - \mu}{\sigma}}}
		- 2\sigma \E_{{\cal N}(0, 1)}\bracket{Y\ind{Y < \frac{\hat{\mu} - \mu}{\sigma}}}
		\\&= \mu - \hat{\mu} + 2(\hat{\mu} - \mu) \paren{\frac{1}{2} + \frac{1}{2}\operatorname{erf}\paren{\frac{\hat{\mu} - \mu}{\sqrt{2}\sigma}}}
		- \frac{\sqrt{2}\sigma}{\sqrt{\pi}} \int_{-\infty}^{\frac{\hat{\mu} - \mu}{\sigma}} y e^{-\frac{y^2}{2}}\diff y
		\\&= (\hat{\mu} - \mu) \operatorname{erf}\paren{\frac{\hat{\mu} - \mu}{\sqrt{2}\sigma}}
		- \frac{\sqrt{2}\sigma}{\sqrt{\pi}} e^{-\frac{(\hat{\mu} - \mu)^2}{2\sigma^2}},
	\end{align*}
	where we used the error function, which is the symmetric function defined for $x\in\R_+$ as
	\[
		\operatorname{erf}(x) = \frac{2}{\sqrt{\pi}} \int_0^x e^{-t^2} \diff t
		= \frac{2}{\sqrt{2 \pi}} \int_0^{\sqrt{2}x} e^{-\frac{u^2}{2}} \diff u
		= 2 \E_{{\cal N}(0, 1)}[\ind{0 \leq Y \leq \sqrt{2} x}].
	\]
	Developing those two functions in the Taylor series leads to the same quadratic behavior.
\end{proof}

Let us now add a context variable.

\begin{lemma}[Reduction to least-squares]
	For $\Y=\R^m$, there exists a $\sigma_m > 0$, such that if $\phi$ is bounded by $\kappa$, and $f^*$ belongs to the class of functions ${\cal F} = \brace{x\to \theta\phi(x)\midvert \theta \in \Y\otimes{\cal H}, \norm{\theta}\leq M}$, and the conditional distribution are distributed as $\paren{Y\midvert X} \sim {\cal N}(f^*(x), \sigma^2 I_m)$, with $\sigma > 2M\kappa\sigma_m$,
	\begin{equation}
		\forall\, f\in{\cal F}, \qquad{\cal R}(f) - {\cal R}(f^*)
		\geq \frac{c_4\norm{f - f^*}_{L^2(\rho_\X)}^2}{2\sigma}.
	\end{equation}
\end{lemma}
\begin{proof}
	According to the precedent lemma, there exists $\sigma_m$ such that $\norm{\hat{\mu} - \mu}\sigma^{-1} \leq \sigma_m^{-1}$ leads to\footnote{This best value for $\sigma_m$ can be derived by studying the Laguerre function, which we will not do in this paper.}
	\[
		\E_{{\cal N}(\mu, \sigma^2 I_m)}[\norm{\hat{\mu} - Y} - \norm{\mu - Y}]
		\geq \frac{c_4\norm{\hat{\mu} - \mu}^2}{2\sigma}.
	\]
	Let $f$ and $f^*\in{\cal F}$ be parametrized by $\theta$ and $\theta^*$.
	For a given $x$, setting $\hat\mu = f_\theta(x) = \theta\phi(x)$ and $\mu = f_{\theta^*}(x)$, we get that, using the operator norm,
	\[
		\norm{\hat{\mu} - \mu} = \norm{(\theta - \theta^*)\phi(x)}
		\leq \norm{\theta - \theta^*}_{\op}\norm{\phi(x)}
		\leq \norm{\theta - \theta^*}\norm{\phi(x)}
		\leq 2M\kappa.
	\]
	Hence, as soon as $2M\kappa \leq \sigma\sigma_m^{-1}$, we have that for almost all $x\in\X$
	\[
		\E_Y\bracket{\norm{f(X) - Y} - \norm{f^*(X) - Y}\midvert X=x}
		\geq \frac{c_4\norm{f(X) - f^*(X)}^2}{2\sigma}.
	\]
	The result follows from integration over $\X$.
\end{proof}

We now have a characterization of the excess of risk that will allow us to reuse lower bounds for least-squares regression.
We will follow the exposition of \cite{Bach2023} that we reproduce and comment here for completeness.
It is based on the generalized Fano's method \citep{Ibragimov1977,Birge1983}.

Learnability over a class of functions depends on the size of this class of functions.
For least-squares regression with a Hilbert class of functions, the right notion of size is given by the Kolmogorov entropy.
Let us call $\epsilon$-packing of ${\cal F}$ with a metric $d$ any family $(f_i)_{i\leq N}\in {\cal F}^N$ such that $d(f_i, f_j) > \epsilon$.
The logarithm of the maximum cardinality of an $\epsilon$-packing defines the $\epsilon$-capacity of the class of functions ${\cal F}$.
We refer the interested reader to Theorem~6 in \cite{Kolmogorov1959} to make a link between the notions of capacity and entropy of a space.
To be perfectly rigorous, the least-squares error in not a norm on the space of $L^2$ functions, but we will call it a {\em quasi-distance} as it verifies symmetry, positive definiteness and the inequality $d(x, y) \leq K(d(x, z) + d(z, y))$ for $K \geq 1$.
Let us define an $\epsilon$-packing with respect to a quasi-distance similarly as before.

The $\epsilon$-capacity of a space ${\cal F}$ gives a lower bound on the number of information to transmit in order to recover a function in ${\cal F}$ up to precision $\epsilon$.
We will leverage this fact in order to show our lower bound.
Let us first reduce the problem to a statistical test.

\begin{lemma}[Reduction to statistical testing]
	Let us consider a class of functions ${\cal F}$ and an $\epsilon$-packing $(f_i)_{i\leq N}$ of ${\cal F}$ with respect to a quasi-distance $d(\cdot, \cdot)$ verifying the triangular inequality up to a multiplicative factor $K$.
	Then the minimax optimality of an algorithm ${\cal A}$ that takes as input the dataset ${\cal D}_n = (X_i, Y_i)_{i \leq n}$ and output a function in ${\cal F}$ can be related to the minimax optimality of an algorithm ${\cal C}$ that takes an input the dataset ${\cal D}_n$ and output an index $j\in[m]$ through
	\begin{equation}
		\inf_{\cal A} \sup_{\rho} \E_{{\cal D}_n\sim\rho^{\otimes n}}\bracket{d\paren{f_{{\cal A}({\cal D}_n)}, f_\rho}}
		\geq \frac{\epsilon}{2K}
		\inf_{\cal C} \sup_{i\in[N]} \Pbb_{{\cal D}_n\sim(\rho_i)^{\otimes n}}\paren{{{\cal C}({\cal D}_n)} \neq i},
	\end{equation}
	where the supremum over $\rho$ has to be understood as taken over all measures whose marginals can be written ${\cal N}(f^*(x), \sigma)$ for $\sigma$ bigger than a threshold $\sigma_m$ and $f^*\in{\cal F}$, and the supremum over $\rho_i$ taken over the same type of measures with $f^* \in (f_i)_{i\leq N}$.
\end{lemma}
\begin{proof}
	Consider an algorithm ${\cal A}$ that takes as input a dataset ${\cal D}_n = (X_j, Y_j)_{j\leq n}$ and output a function $f\in{\cal F}$.
	We would like to see ${\cal A}$ as deriving from a classification rule and relate the classification and regression errors.
	The natural classification rule associated with the algorithm ${\cal A}$ can be defined through $\pi$ the projection from ${\cal F}$ to $[N]$ that minimizes $d\paren{f, f_{\pi(f)}}$.
	The classification error and regression error made by $\pi\circ{\cal A}$ can be related thanks to the $\epsilon$-packing property.
	For any index $j\in[N]$
	\[
		d\paren{f_{\pi\circ{\cal A}({\cal D}_n)}, f_j} \geq \epsilon \ind{\pi\circ{\cal A}({\cal D}_n) \neq j}.
	\]
	The error made by $f_{\cal A}({\cal D}_n)$ relates to the one made by $f_{\pi\circ{\cal A}({\cal D}_n)}$ thanks to the modified triangular inequality, using the definition of the projection
	\[
		d\paren{f_{\pi\circ{\cal A}({\cal D}_n)}, f_j}
		\leq K \paren{d\paren{f_{\pi\circ{\cal A}({\cal D}_n)}, f_{{\cal A}({\cal D}_n)}}
			+ d\paren{f_{{\cal A}({\cal D}_n)}, f_j}}
		\leq 2 Kd\paren{f_{{\cal A}({\cal D}_n)}, f_j}.
	\]
	Finally,
	\[
		d\paren{f_{{\cal A}({\cal D}_n)}, f_{j}} \geq \frac{\epsilon}{2K} \ind{\pi\circ{\cal A}({\cal D}_n) \neq j}.
	\]
	Assuming that the data were generated by a $\rho_i$ and taking the expectation, the supremum over $\rho_i$ and the infimum over ${\cal A}$ leads to
	\[
		\inf_{\cal A}\sup_{\rho_i} \E_{{\cal D}_n\sim\rho_i^{\otimes n}}\bracket{d\paren{f_{{\cal A}({\cal D}_n)}, f_i}}
		\geq \frac{\epsilon}{2K}\inf_{{\cal C}=\pi\circ{\cal A}}\sup_{(\rho_i)} \Pbb_{{\cal D}_n\sim\rho_i^{\otimes n}}\paren{{\cal C}({\cal D}_n) \neq i}.
	\]
	Because $\pi\circ{\cal A}$ are part of classification rules (indeed it parametrizes all the classification rules, simply consider ${\cal A}$ that matches a dataset to one of the functions $(f_i)_{i\leq N}$), and because the distributions $\rho_i$ are part of the distributions $\rho$ defined in the lemma, this last equation implies the stated result.
\end{proof}

One of the harshest inequalities in the last proof is due to the usage of the $\epsilon$-packing condition without considering error made by $d\paren{f_{\pi\circ{\cal A}({\cal D}_n)}, f_j}$ that might be much worse than $\epsilon$.
We will later add a condition on the $\epsilon$-packings to ensure that the $(f_i)$ are not too far from each other.
This will not be a major problem when considering small balls in big dimension spaces.

\subsubsection{Results from statistical testing}
In this section, we expand on lower bounds for statistical testing.
We refer the curious reader to \cite{Cover1991}.
We begin by relaxing the supremum by an average
\begin{align}
	\inf_{\cal C} \sup_{i\in[N]} \Pbb_{{\cal D}_n\sim(\rho_i)^{\otimes n}}\paren{{{\cal C}({\cal D}_n)} \neq i}
	 & = \inf_{\cal C} \sup_{p\in\prob{N}} \sum_{i=1}^N p_i \Pbb_{{\cal D}_n\sim(\rho_i)^{\otimes n}}\paren{{{\cal C}({\cal D}_n)} \neq i}
	\\&\geq \inf_{\cal C} \frac{1}{N}\sum_{i=1}^N \Pbb_{{\cal D}_n\sim(\rho_i)^{\otimes n}}\paren{{{\cal C}({\cal D}_n)} \neq i}.
\end{align}
The last quantity can be seen as the best measure of error that can be achieved by a decoder ${\cal C}$ of a signal $i\in[N]$ based on noisy observations ${\cal D}_n$ of the signal.
A lower bound on such a similar quantity is the object of Fano's inequality \citep{Fano1968}.

\begin{lemma}[Fano's inequality]
	Let $(X, Y)$ be a couple of random variables in $\X\times\Y$ with $\X$, $\Y$ finite, and $\hat{X}:\Y\to\X$ be a classification rule.
	Then, the error $e = e(X, Y) = \ind{X \neq \hat{X}(Y)}$ verifies
	\[
		H\paren{X\midvert Y} \leq H(e) + \Pbb(e) \log(\card{\X} - 1) \leq \log(2) + \Pbb(e)\log(\card\X).
	\]
	Where for $(X, Y)\in\prob{\X\times\Y}$, $H(X)$ and $H\paren{X\midvert Y}$ denotes the entropy and conditional entropy, defined as, with the convention $0\log 0 = 0$,
	\begin{align*}
		 & H(X) = -\sum_{x\in\X} \Pbb(X=x) \log(\Pbb(X=x)),
		\\ &H\paren{X\midvert Y} = -\sum_{x\in\X,y\in\Y} \Pbb(X=x, Y=y) \log(\Pbb\paren{X=x\midvert Y=y}).
	\end{align*}
\end{lemma}

\begin{proof}
	This lemma is actually the result of two properties.
	The first part of the proof is due to some manipulation of the entropy, consisting in showing that
	\begin{equation}
		\label{sgd:eq:proc_1}
		H\paren{X\midvert \hat{X}(Y)} \leq H(e) + \Pbb(e)\log(\card\X - 1).
	\end{equation}
	Let us first recall the following additive property of entropy
	\begin{align*}
		H\paren{X, Y\midvert Z} & = -\sum_{x\in\X, y\in\Y, z\in{\cal Z}} \Pbb(X=x, Y=y, Z=z)\log(\Pbb\paren{X=x,Y=y\midvert Z=z})
		\\&= -\sum_{x\in\X, y\in\Y, z\in{\cal Z}} \Pbb(X=x, Y=y, Z=z)\log(\Pbb\paren{Y=y\midvert X=x, Z=z})
		\\&\qquad\qquad\qquad\qquad - \sum_{x\in\X, y\in\Y, z\in{\cal Z}} \Pbb(X=x, Y=y, Z=z)\log(\Pbb\paren{X=x\midvert Z=z})
		\\&= H\paren{Y\midvert X, Z} + H\paren{X\midvert Z}.
	\end{align*}
	Using this chain rule, we get
	\begin{align*}
		H\paren{e, X\midvert \hat{X}} & = H\paren{e\midvert X, \hat{X}} + H\paren{X\midvert\hat{X}}
		\\&= H\paren{X\midvert e, \hat{X}} + H\paren{e\midvert\hat{X}}
	\end{align*}
	Because $e$ is a function of $\hat{X}$ and $X$ one can check that $H\paren{e\midvert X, \hat{X}} = 0$,
	\begin{align*}
		H\paren{e\midvert X,\hat{X}}
		 & = -\sum_{e, X, \hat{X}} \Pbb(X, \hat{X}) \Pbb\paren{e\midvert X,\hat{X}} \log(\Pbb\paren{e\midvert X,\hat{X}})
		\\&= -\sum_{e, X, \hat{X}} \Pbb(X, \hat{X}) \ind{e=\ind{X\neq\hat{X}}}\log(\ind{e=\ind{X\neq\hat{X}}})
		= -\sum_{e, X,\hat{X}} \Pbb(X, \hat{X}) \cdot 0 = 0.
	\end{align*}
	Using Jensen inequality for the logarithm, we get
	\begin{align*}
		H\paren{X\midvert e, \hat{X}}
		 & = -\sum_{X, e, \hat{X}} \Pbb(X, e, \hat{X}) \log(\Pbb\paren{X\midvert e, \hat{X}})
		\\&= -\sum_{x, x'} \Pbb(X=x, e=0, \hat{X}=x') \log(\Pbb\paren{X=x\midvert e=0, \hat{X}=x'})
		\\&\qquad\qquad\qquad\qquad- \Pbb(X=x, e=1, \hat{X}=x') \log(\Pbb\paren{X=x\midvert e=1, \hat{X}=x'})
		\\&= -\sum_{x, x'} \Pbb\paren{X = x, \hat{X}=x'}\ind{x=x'} \log(\ind{x=x'})
		\\&\qquad\qquad\qquad\qquad- \Pbb(e=1)\ind{x\neq x'}\Pbb(X=x, \hat{X}=x') \log(\Pbb\paren{X=x\midvert \hat{X}=x'})
		\\&= \Pbb(e=1)\sum_{x'} \Pbb(\hat{X}=x') \sum_{x\neq x'} \Pbb\paren{X=x\midvert \hat{X}=x'} \log\paren{\frac{1}{\Pbb\paren{X=x\midvert \hat{X}=x'}}}
		\\&\leq \Pbb(e=1)\sum_{x'} \Pbb(\hat{X}=x') \log\paren{\sum_{x\neq x'} \Pbb\paren{X=x\midvert \hat{X}=x'} \frac{1}{\Pbb\paren{X=x\midvert \hat{X}=x'}}}
		\\&= \Pbb(e=1)\log(\card{\X} - 1).
	\end{align*}
	Using that conditioning reduces the entropy, which follows again from Jensen inequality,
	\begin{align*}
		H(X) - H\paren{X\midvert Y}
		 & = \sum_{x, y} \Pbb(X=x,Y=y) \log\paren{\frac{\Pbb\paren{X=x\midvert Y=y}}{\Pbb\paren{X=x}}}
		\\&= -\sum_{x, y} \Pbb(X=x,Y=y) \log\paren{\frac{\Pbb\paren{X=x}\Pbb\paren{Y=y}}{\Pbb\paren{X=x, Y=y}}}
		\\&\geq -\log\paren{\sum_{x, y} \Pbb(X=x,Y=y) \frac{\Pbb\paren{X=x}\Pbb\paren{Y=y}}{\Pbb\paren{X=x, Y=y}}} = 0,
	\end{align*}
	we get
	\[
		H\paren{e \midvert \hat{X}} \leq H(e) \leq \log(2).
	\]
	Hence, we have proven that
	\[
		H\paren{X\midvert \hat{X}} \leq \Pbb(e=1)\log(\card\X - 1) + H(e).
	\]

	The rest of the proof follows from the so-called data processing inequality, that is
	\begin{equation}
		H\paren{X\midvert \hat{X}(Y)} \geq H\paren{X\midvert Y}.
	\end{equation}
	We will not derive it here, since it will not be used in the following.
\end{proof}

In our case, a slight modification of the proof of Fano's inequality leads to the following Proposition.
\begin{lemma}[Generalized Fano's method]
	For any family of distributions $(\rho_i)_{i\leq N}$ on $\X\times\Y$ with $N\in\N^*$, any classification rule ${\cal C}:{\cal D}_n \to [N]$ cannot beat the following average lower bound
	\begin{equation}
		\inf_{\cal C} \frac{1}{N}\sum_{i=1}^N \Pbb_{{\cal D}_n \sim \rho_i^{\otimes n}}({\cal C}({\cal D}_n) \neq i) \log(N - 1) \geq \log(N) - \log(2) - \frac{n}{N^2} \sum_{i, j \in [N]} K\paren{\rho_i\midvertvert \rho_j},
	\end{equation}
	where $K\paren{p\midvertvert q}$ is the Kullback-Leibler divergence defined for any measure $p$ absolutely continuous with respect to a measure $q$ as
	\[
		K\paren{p\midvertvert q} = \E_{X\sim q}\bracket{-\log\paren{\frac{\diff p(X)}{\diff q(X)}}}.
	\]
\end{lemma}
\begin{proof}
	Let us consider the joint variable $(X, Y)$ where $X$ is a uniform variable on $[N]$ and $\paren{Y\midvert X}$ is distributed according to $\rho_X^{\otimes n}$.
	For any classification rule $\hat{X}:{\cal D}_n\to[N]$, using~\eqref{sgd:eq:proc_1} we get
	\[
		\frac{1}{N}\sum_{i=1}^N \Pbb_{{\cal D}_n\sim\rho_i^{\otimes n}}\paren{\hat{X}({\cal D}_n) \neq i} = \Pbb(\hat{X} \neq X)\log(N-1) \geq H\paren{X\midvert \hat{X}} - \log(2).
	\]

	We should work on $H\paren{X\midvert \hat{X}\midvert X}$ with similar ideas to the data processing inequality.
	First of all, using the chain rule for entropy
	\[
		H\paren{X\midvert \hat{X}} = H(X, \hat{X}) - H(\hat{X}) = H(X) + (H(X, \hat{X}) - H(X) - H(\hat{X})) = \log(N) - I(X,\hat{X}),
	\]
	where $I$ is the mutual information defined as, for $X$ and $Z$ discrete
	\begin{align*}
		I(X, Z) & = H(X) + H(Z) - H(X, Z)
		= \sum_{x, z} \Pbb\paren{X=x, Z=z} \log\paren{\frac{\Pbb(X=x, Z=z)}{\Pbb(X=x)\Pbb(Z=z)}}
		\\&= \sum_x \Pbb(X=x) \sum_z \Pbb\paren{Z=z\midvert X=x} \log\paren{\frac{\Pbb\paren{Z=z\midvert X=x)}}{\Pbb(Z=z)}}.
	\end{align*}
	Similarly, one can define the mutual information for continuous variables.
	In particular, we are interested in the case where $X$ is discrete and $Y$ is continuous, denote by $\mu_\Y$ the marginal of $(X, Y)$ over $Y$ and by $\mu\vert_x$ the conditional $\paren{Y\midvert X=x}$.
	\[
		I(X, Y) = \sum_{x}\Pbb(X=x)\int_{y} \mu\vert_x(\diff y) \log\paren{\frac{\mu\vert_x(\diff y)}{\mu(\diff y)}}.
	\]

	Let us show the following version of the data processing inequality
	\begin{equation}
		I(X, \hat{X}(Y)) \leq I(X, Y).
	\end{equation}
	To do so, we will use the conditional independence of $X$ and $\hat{X}$ given $Y$, which leads to
	\begin{align*}
		\Pbb\paren{X=x\midvert \hat{X}=x'}
		 & = \int \Pbb\paren{X=x\midvert Y=\diff y}\Pbb\paren{Y=\diff y\midvert \hat{X}=z}
		\\&= \int \frac{\Pbb(X=x)\mu\vert_x(\diff y)}{\mu(\diff y)}\Pbb\paren{Y=\diff y\midvert \hat{X}=z}.
	\end{align*}
	Hence, using Jensen inequality,
	\begin{align*}
		I(X, \hat{X})
		 & = H(X) - H\paren{X\midvert \hat{X}}
		\\&= H(X) + \sum_{z} \Pbb(\hat{X}=z) \sum_{x} \Pbb(X=x) \log(\Pbb\paren{X=x\midvert \hat{X}=z})
		\\&= H(X) + \sum_{z} \Pbb(\hat{X}=z) \sum_{x} \Pbb(X=x) \log\paren{\int \frac{\Pbb(X=x)\mu\vert_x(\diff y)}{\mu(\diff y)}\Pbb\paren{Y=\diff y\midvert \hat{X}=z}}
		\\&\leq H(X) + \sum_{z} \Pbb(\hat{X}=z) \sum_{x} \Pbb(X=x)\int \Pbb\paren{Y=\diff y\midvert \hat{X}=z} \log\paren{\frac{\Pbb(X=x)\mu\vert_x(\diff y)}{\mu(\diff y)}}
		\\&= H(X) + \sum_{x} \Pbb(X=x) \int \mu(\diff y) \log\paren{\frac{\Pbb(X=x)\mu\vert_x(\diff y)}{\mu(\diff y)}}
		\\&=\sum_{x} \Pbb(X=x) \paren{\int \mu(\diff y) \log\paren{\frac{\Pbb(X=x)\mu\vert_x(\diff y)}{\mu(\diff y)}} - \log(P(X=x)}
		\\&=\sum_{x} \Pbb(X=x) \int \mu(\diff y) \log\paren{\frac{\mu\vert_x(\diff y)}{\mu(\diff y)}}
		\\&= I(X, Y).
	\end{align*}

	We continue by computing the value of $I(X, Y)$, by definition and using Jensen inequality, we get
	\begin{align*}
		I(X, Y) & = \frac{1}{N}\sum_{i\in[N]} \int_{{\cal D}_n\sim\rho_i^{\otimes n}} \rho_i^{\otimes n}(\diff {\cal D}_n) \log\paren{\frac{\rho_i^{\otimes n}(\diff {\cal D}_n)}{\frac{1}{N}\sum_{j\in[N]} \rho_j^{\otimes n}(\diff {\cal D}_n)}}
		\\&\leq \frac{1}{N}\sum_{i\in[N]} \int_{{\cal D}_n\sim\rho_i^{\otimes n}} \rho_i^{\otimes n}(\diff {\cal D}_n) \frac{1}{N}\sum_{j\in[N]}\log\paren{\frac{\rho_i^{\otimes n}(\diff {\cal D}_n)}{\rho_j^{\otimes n}(\diff {\cal D}_n)}}
		= \frac{1}{N^2}\sum_{i,j \in[N]} K\paren{\rho_i^{\otimes n}\midvertvert\rho_j^{\otimes n}}.
	\end{align*}
	We conclude from the fact that for $p$ and $q$ two distributions on a space ${\cal Z}$, we have
	\begin{align*}
		K\paren{p^{\otimes n}\midvertvert q^{\otimes n}}
		 & = \int_{{\cal Z}^n} -\log\paren{\frac{\diff p^{\otimes n}(z_1,\cdots, z_n)}{\diff q^{\otimes n}(z_1,\cdots, z_n)}} q^{\otimes n}(\diff z_1,\cdots, \diff z_n)
		\\&= \int_{{\cal Z}^n} -\log\paren{\frac{\prod_{i\leq n}\diff p(z_i)}{\prod_{i\leq n}\diff q(z_i)}} q^{\otimes n}(\diff z_1,\cdots, \diff z_n)
		\\&= \sum_{i\leq n}\int_{{\cal Z}^n} -\log\paren{\frac{\diff p(z_i)}{\diff q(z_i)}} q^{\otimes n}(\diff z_1,\cdots, \diff z_n)
		\\&= \sum_{i\leq n}\int_{\cal Z} -\log\paren{\frac{\diff p(z_i)}{\diff q(z_i)}} q(\diff z_i)
		= n K\paren{p\midvertvert q}.
	\end{align*}
	This explains the result.
\end{proof}

Let us assemble all the results proven thus far.
In order to reduce our excess risk to a quadratic metric, we have assumed that the conditional distribution $\rho_i\vert_x$ to be Gaussian noise.
In order to integrate this constraint into the precedent derivations, we leverage the following lemma.

\begin{lemma}[Kullback-Leibler divergence with Gaussian noise]
	If $\rho_i$ and $\rho_j$ are two different distributions on $\X\times\Y$ such that there marginal over $\X$ are equal and the conditional distributions $\paren{Y\midvert X=x}$ are respectively equal to ${\cal N}(f_i(x), \sigma I_m)$ and ${\cal N}(f_j(x), \sigma I_m)$, then
	\[
		K\paren{\rho_i \midvertvert \rho_j} = \frac{1}{2\sigma^2} \norm{f_i - f_j}^2_{L^2(\rho_\X)}.
	\]
\end{lemma}
\begin{proof}
	We proceed with
	\begin{align*}
		K\paren{\rho_i \midvertvert \rho_j}
		 & = \int_\X \E_{Y\sim{\cal N}(f_j(x), \sigma I_m)}\bracket{\frac{\norm{Y-f_i(x)}^2 - \norm{Y-f_j(x)}^2}{2\sigma^2}}\rho_j(\diff x)
		\\&= \int_\X \E_{Y\sim{\cal N}\paren{\frac{f_j(x) - f_i(x)}{\sqrt{2}\sigma}, I_m}}\bracket{\norm{Y}^2} - \E_{Y\sim{\cal N}(0, I_m)}\bracket{\norm{Y}^2}\rho_j(\diff x)
		\\&= \int_\X \paren{m + \frac{\norm{f_j(x) - f_i(x)}^2}{2\sigma^2} - m} \rho_j(\diff x)
		= \frac{\norm{f_j - f_i}^2_{L^2(\rho_\X)}}{2\sigma^2},
	\end{align*}
	where we have used the fact that the mean of a non-central $\chi$-square variable of parameter $(m, \mu^2)$ is $m+\mu^2$.
	One could also develop the first two squared norms and use the fact that for any vector $u\in\R^m$, $\E[\scap{Y-f_i(x)}{u}] = 0$ to get the result.
\end{proof}

Combining the different results leads to the following proposition.
\begin{lemma}
	\label{sgd:prop:step_1}
	Under Assumption \ref{sgd:ass:source} with ${\cal F} = \brace{x\in \X\to \theta\phi(x) \in \Y\midvert \norm{\theta} \leq M}$ and $\phi$ bounded by $\kappa$, for any family $(f_i)_{i\leq N_\epsilon} \in {\cal F}^N$ and any $\sigma > 2M\kappa \sigma_m$
	\begin{align*}
		 & \inf_{\cal A}\sup_\rho \E_{{\cal D}_n\sim\rho^{\otimes n}}[{\cal R}(f_{{\cal A}({\cal D}_n)}; \rho)] - {\cal R}^*(\rho)
		\\&\qquad\qquad\qquad\geq \frac{\min_{i,j\in[N]} \norm{f_i - f_j}_{L^2(\rho_\X)}^2}{16(m+2)^{1/2}\sigma} \paren{1 - \frac{\log(2)}{\log(N)} - \frac{n\max_{i,j\in[N]} \norm{f_i - f_j}^2_{L^2(\rho_\X)}}{2\sigma^2 \log(N)}},
	\end{align*}
	for any algorithm ${\cal A}$ that maps a dataset ${\cal D}_n \in (\X\times\Y)^n$ to a parameter $\theta \in \Theta$.
\end{lemma}

\subsubsection{Covering number for linear model}

We are left with finding a good packing of the space induced by Assumption \ref{sgd:ass:source}.
To do so, we shall recall some property of reproducing kernel methods.

\begin{lemma}[Linear models are ellipsoids]
	\label{sgd:lem:lin_ell}
	For ${\cal H}$ a separable Hilbert space and $\phi:\X\to{\cal H}$ bounded, the class of functions ${\cal F} = \brace{x\in \X\to \theta\phi(x) \in \Y\midvert \norm{\theta} \leq M}$ can be characterized by
	\begin{equation}
		{\cal F} = \brace{f:\X\to\Y\midvert \norm{K^{-1/2}f}_{L^2(\rho_\X)} \leq M},
	\end{equation}
	where $\rho_\X$ is any distribution on $\X$ and $K$ is the operator on $L^2(\rho_\X)$ that map $f$ to
	\[
		Kf(x') = \int_{x\in\X} \scap{\phi(x)}{\phi(x')} f(x) \rho_\X(\diff x),
	\]
	whose image is assumed to be dense in $L^2$.
\end{lemma}
\begin{proof}
	This follows for isometry between elements in ${\cal H}$ and elements in $L^2$.
	More precisely, let us define
	\[\myfunction{S}{\Y\otimes{\cal H}}{L^2(\X, \Y, \rho_\X)}{\theta}{x\to \theta\phi(x).}\]
	The adjoint of $S$ is characterized by
	\[\myfunction{S^*}{L^2(\X, \Y, \rho_\X)}{\Y\otimes{\cal H}}{f}{\E[f(x)\otimes\phi(X)],}\]
	which follows from the fact that for $\theta \in \Y\otimes{\cal H}$, $f\in L^2$ we have
	\begin{align*}
		\scap{\theta}{S^*f}_{\Y\otimes{\cal H}} = \scap{S\theta}{f}_{L^2} & = \sum_{i=1}^m \int_{\X} f_i(x) \scap{\theta_i}{\phi(x)}_{\cal H} \rho_\X(\diff x)
		\\&= \sum_{i=1}^m \scap{\theta_i}{\E[f_i(X) \phi(X)]}_{\cal H}
		= \scap{\theta}{\E[f(X)\otimes\phi(X)]}_{\Y\otimes{\cal H}}.
	\end{align*}
	When $SS^*$ is compact and dense in $L^2$, we have
	\[
		\norm{\theta}_{\Y\otimes{\cal H}} = \norm{(SS^*)^{-1/2} S\theta}_{L^2(\rho_\X)}.
	\]
	The compactness allows considering spectral decomposition hence fractional powers.
	We continue by observing that $SS^* = K$, which follows from
	\begin{align*}
		(SS^*f)(x') = (S\E[f(X)\otimes\phi(X)])(x')
		= \E[f(X)\otimes\phi(X)] \phi(x')
		= \E[\scap{\phi(X)}{\phi(x')} f(X)].
	\end{align*}
	The compactness of $K$ derives from the fact that
	\[
		\norm{Kf(x')}^2 = \norm{\E[\scap{\phi(X)}{\phi(x')} f(X)]}^2
		\leq \E[\norm{\scap{\phi(X)}{\phi(x')} f(X)}^2]
		\leq \kappa^2 \norm{f}_{L^2}^2.
	\]
	Hence, $\norm{K}_{\text{op}} \leq \kappa^2$.
	Indeed, it is not hard to prove that the trace of $K$ is bounded by $m\kappa^2$, hence $K$ is not only compact but trace-class.
\end{proof}

It should be noted that the condition on $K$ being dense in $L^2(\rho_\X)$ is not restrictive, as indeed all the problem is only seen through the lens of $\phi$ and $\rho_\X$: one can replace $\X$ by $\supp\rho_\X$ and $L^2(\rho_\X)$ by the closure of the range of $K$ in $L^2(\rho_\X)$ without modifying nor the analysis, nor the original problem.

We should study packing in the ellipsoid ${\cal F} = \brace{f\in L^2\midvert \|K^{-1/2}f\|_{L^2(\rho_\X)} \leq M}$.
It is useful to split the ellipsoid between a projection on a finite dimensional space that is isomorphic to the Euclidean space $\R^k$ and on a residual space $R$ where the energies $(\norm{f\vert_R}_{L^2(\rho_\X)})_{f\in{\cal F}}^2$ are uniformly small.
We begin with the following packing lemma, sometimes referred to as Gilbert-Varshamov bound \citep{Gilbert1952,Varshamov1957} which corresponds to a more generic result in coding theory.

\begin{lemma}[$\ell_2^2$-packing of the hypercube]
	For any $k \in \N^*$, there exists a $k/4$-packing of the hypercube $\brace{0, 1}^k$, with respect to Hamming distance, of cardinality $N = \exp(k/8)$.
\end{lemma}
\begin{proof}
	Let us consider $\epsilon > 0$, and a maximal $\epsilon$-packing $(x_i)_{i\leq N}$ of the hypercube with respect to the distance $d(x, y) = \sum_{i\in[k]} \ind{x_i\neq y_i} = \norm{x-y}_1 = \norm{x-y}_2^2$.
	By maximality, we have $\brace{0,1}^k \subset \cup_{i\in[N]}B_d(x_i,\epsilon)$, hence
	\[
		2^k \leq N \card{\brace{x\in\brace{0,1}^k\midvert \norm{x}_1 \leq \epsilon}}.
	\]
	This inequality can be rewritten with $Z$ a binomial variable of parameter $(k, \sfrac{1}{2})$ as
	\(
	1 \leq N \Pbb(Z \leq \epsilon).
	\)
	Using Hoeffding inequality \citep{Hoeffding1963}, when $\epsilon = k/4$ we get
	\[
		N^{-1} \leq \Pbb(Z \leq k/4) = \Pbb(Z - \E[Z] \leq k/4)
		\leq \exp\paren{-\frac{2k^2}{4^2 k}} = \exp\paren{-k/8}.
	\]
	This is the desired result.
\end{proof}

\begin{lemma}[Packing of infinite-dimensional ellipsoids]
	\label{sgd:lem:pack_ell}
	Let ${\cal F}$ be the function in $L^2(\rho_\X)$ such that $\norm{K^{-1/2}f}_{L^2(\rho_\X)}\leq M$ for $K$ a compact operator and $M$ any positive number.
	For any $k\in\N^*$, it is possible to find a family of $N \geq \exp(k/8)$ elements $(f_i)_{i\in[N]}$ in ${\cal F}$ such that for any $i\neq j$,
	\begin{equation}
		\frac{kM^2}{\sum_{i\leq k} \lambda_i^{-1}} \leq \norm{f_i - f_j}_{L^2(\rho_\X)}^2 \leq \frac{4k M^2}{\sum_{i\leq k}\lambda_i^{-1}},
	\end{equation}
	where $(\lambda_i)_{i\in\N}$ are the ordered (with repetition) eigenvalues of $K$.
\end{lemma}
\begin{proof}
	Let us denote by $(\lambda_i)_{i\in \N}$ the eigenvalues of $K$ and $(u_i)_{i\in\N}$ in $L^2$ the associated eigenvectors.
	Consider $(a_s)_{s\in[N]}$ a $k$-packing of the hypercube $\brace{-1,1}^k$ for $N \geq \exp(k/8)$ with respect to the $\ell^2_2$ quasi-distance and define for any $a\in\brace{a_s}$
	\[
		f_a = \frac{M}{c} \sum_{s=1}^k a_i u_i,
	\]
	with $c^2 = \sum_{i=1}^k \lambda_i^{-1}$.
	We verify that
	\begin{align*}
		 & \norm{K^{-1/2} f_a}_{L^2}^2 = \frac{M^2}{c^2} \sum_{i=1}^k \lambda_i^{-1} = M^2.
		\\&\norm{f_a - f_b}_{L^2}^2 = \frac{M^2}{c^2} \sum_{i=1}^k \abs{a_i - b_i}^2
		= \frac{M^2}{c^2} \norm{a_i - b_i}_2^2
		\in \frac{M^2}{c^2}\cdot [k, 4k].
	\end{align*}
	This is the object of the lemma.
\end{proof}

So far, we have proven the following lower bound.
\begin{lemma}
	\label{sgd:prop:step_2}
	Under Assumption \ref{sgd:ass:source} with ${\cal F} = \brace{x\in \X\to \theta\phi(x) \in \Y\midvert \norm{\theta} \leq M}$ and $\phi$ bounded by $\kappa$, for any family $(f_i)_{i\leq N_\epsilon} \in {\cal F}^N$ and any $\sigma > 2M\kappa \sigma_m$ and $km > 10$,
	\[
		\inf_{\cal A}\sup_\rho \E_{{\cal D}_n\sim\rho^{\otimes n}}[{\cal R}(f_{{\cal A}({\cal D}_n)}; \rho)] - {\cal R}^*(\rho)
		\geq \frac{1}{128}\min\brace{\frac{M^2}{\sigma m^{1/2}\sum_{i\leq k} (k\lambda_i)^{-1}}, \frac{\sigma km^{1/2}}{32 n}},
	\]
	for any algorithm ${\cal A}$ that maps a dataset ${\cal D}_n \in (\X\times\Y)^n$ to a parameter $\theta \in \Theta$, and where $(\lambda_i)$ are the ordered eigenvalue of the operator $K$ on $L^2(\X,\R,\rho_\X)$ that maps any function $f$ to the function $Kf$ defines for $x'\in\X$ as
	\[
		(Kf)(x') = \int_{x\in\X} \scap{\phi(x)}{\phi(x')} f(x)\rho_\X(\diff x).
	\]
	In particular, when $\lambda_i = \kappa^2 i^{-a} / \zeta(\alpha)$, where $\zeta$ denotes the Riemann zeta function, we get the following bounds.
	If we optimize with respect to $\sigma$, there exists $n_\alpha \in \N$ such that for any $n > n_\alpha$.
	\begin{equation}
		\inf_{\cal A}\sup_\rho \E_{{\cal D}_n\sim\rho^{\otimes n}}[{\cal R}(f_{{\cal A}({\cal D}_n)}; \rho)] - {\cal R}^*(\rho)
		\geq \frac{M\kappa}{725\zeta(\alpha)^{1/2} n^{1/2}}.
	\end{equation}
	If we fix $\sigma = \beta M\kappa$ with $\beta \geq 2$, and we optimize with respect to $k$, there exists a constant $c_\beta$ and an integer $n_0$ such that for $n > n_0$ we have
	\begin{equation}
		\inf_{\cal A}\sup_\rho \E_{{\cal D}_n\sim\rho^{\otimes n}}[{\cal R}(f_{{\cal A}({\cal D}_n)}; \rho)] - {\cal R}^*(\rho)
		\geq \frac{M\kappa c_\beta}{\zeta(\alpha)^{\frac{1}{1+\alpha}} n^{\frac{\alpha}{\alpha+1}}}.
	\end{equation}
\end{lemma}
\begin{proof}
	Reusing Lemma \ref{sgd:prop:step_1}, with the same notations, we have the lower bound in
	\[
		\frac{\min\norm{f_i - f_j}^2}{16\sigma(m+2)^{1/2}}\paren{1 - \frac{\log(2)}{\log(N)} - \frac{n\max\norm{f_i - f_j}^2}{2\sigma^2\log(N)}}.
	\]
	Let $K$ and $K_\Y$ be the self-adjoint operators on $L^2(\X, \R, \rho_\X)$ and $L^2(\X, \Y, \rho_\X)$ respectively, both defined through the formula
	\[
		(Kf)(x') = \int_{x\in\X} \scap{\phi(x)}{\phi(x')} f(x)\rho_\X(\diff x).
	\]
	When $K$ is compact, it admits an eigenvalue decomposition $K = \sum_{i\in\N} \lambda_i u_i\otimes u_i$ where the equality as to be understood as the convergence of operator with respect to the operator norm based on the $L^2$-topology.
	It follows from the product structure of $L^2(\X, \Y, \rho_\X) \simeq L^2(\X,\R,\rho_\X)^m$ that $K_\Y = \sum_{i\in\R, j\in[m]} \sum_{i\in\N, j\in[m]} \lambda_i (e_i\otimes y_j) \otimes (e_i\otimes u_j)$ with $(e_j)$ the canonical basis of $\Y = \R^m$.
	As a consequence, if $(\lambda_i)_{i\in\N}$ are the ordered eigenvalues of $K$ then $(\lambda_{\floor{i / m}})$ are the ordered eigenvalues of $K_\Y$.
	Hence, with Lemmas \ref{sgd:lem:lin_ell} and \ref{sgd:lem:pack_ell}, it is possible to find $N=\exp(km/8)$ functions in ${\cal F}$ such that
	\[
		\frac{km M^2}{m\sum_{i\leq k} \lambda_i^{-1}} \leq \norm{f_i - f_j}_{L^2(\rho_\X)}^2 \leq \frac{4km M^2}{m\sum_{i\leq k}\lambda_i^{-1}}.
	\]
	If we multiply those functions by $\eta\in[0,1]$ we get a lower bound in
	\[
		\frac{\eta^2 M^2}{16\sigma(m+2)^{1/2}\sum_{i\leq k}(k\lambda_i)^{-1}}\paren{1 - \frac{8\log(2)}{km} - \frac{16 M^2 n\eta^2}{\sigma^2 km \sum_{i\leq k} (k\lambda_i)^{-1}}}.
	\]
	Making sure that the last two terms are smaller than one fourth and one half respectively we get the following conditions on $k$ and $\eta$, with $\Lambda_k = \sum_{i\leq k} (k\lambda_i)^{-1}$,
	\[
		km \geq 32 \log(2),\qquad
		32M^2 n \eta^2 \leq \sigma^2 km\Lambda_k.
	\]
	Using the fact that $\eta < 1$, the lower bound becomes
	\[
		\frac{M^2}{128\sigma m^{1/2}\Lambda_k} \min\brace{1, \frac{\sigma^2 km \Lambda_k}{32M^2 n}}
		=\frac{1}{128}\min\brace{\frac{M^2}{\sigma m^{1/2}\Lambda_k}, \frac{\sigma km^{1/2}}{32 n}},
	\]
	as long as $km > 10$.
	When $\lambda_i^{-1} = i^\alpha \zeta(\alpha) / \kappa^2$, since $\Lambda_k \leq \lambda_k^{-1}$, we simplify the last expression as
	\[
		\frac{1}{128}\min\brace{\frac{M^2 \kappa^2}{\sigma m^{1/2}k^\alpha \zeta(\alpha)}, \frac{\sigma km^{1/2}}{32 n}}.
	\]
	Optimizing with respect to $\sigma$ leads to
	\[
		\sigma^2 = \frac{32nM^2\kappa^2}{mk^{1+\alpha} \zeta(\alpha)} \geq 4M^2\kappa^2 \sigma_m.
	\]
	This gives
	\[
		n_{\alpha, m} = m\zeta(\alpha) \sigma_m^2 / 8.
	\]
	The dependency of $n_\alpha$ to $m$ can be removed since any problem with $\Y=\R^m$ can be cast as a problem in $\R^{m+1}$ by adding a spurious coordinate.
	Taking $k=1$ and $m=10$ leads to the result stated in the lemma.
	When $n < n_\alpha$, one can artificially multiply the bound by $n_\alpha^{1/2}$, since an optimal algorithm can not do better with fewer data.
	After checking that one can take $\sigma_1 \geq 1$, this leads to a bound in
	\[
		\frac{M\kappa}{2048 n^{1/2}}.
	\]
	Optimizing with respect to $k$ leads to $k^{\alpha + 1} = 32 M^2 \kappa^2 n / (\sigma^2 m \zeta(\alpha))$ and a bound in
	\[
		\frac{(\sigma m^{1/2})^{\frac{\alpha-1}{\alpha+1}} (M\kappa)^{\frac{2}{\alpha+1}}}{128(32 n)^{\frac{\alpha}{\alpha+1}} \zeta(\alpha)^{\frac{1}{\alpha+1}}}.
	\]
	The condition $k > \min\brace{10m^{-1}, 1}$ and $\sigma \geq 2M\kappa\sigma_m$ translates into the condition
	\[
		4M^2\kappa^2\sigma_m^2 \leq \sigma^2 \leq \frac{32 M^2\kappa^2 n}{m\zeta(\alpha)} \min\brace{1, \frac{m^{1+\alpha}}{10^{1+\alpha}}}.
	\]
	We deduce that $\sigma_m = O(m^{-1/2})$, otherwise we would not respect the upper bound derived with Rademacher complexity (or have made a mistake somewhere).
	Once again we can remove the dependency to $m$. Considering $\sigma =\beta M\kappa$ leads to the result stated in the lemma.
\end{proof}

\subsubsection{Controlling eigenvalues decay}

Based on Lemma \ref{sgd:prop:step_2}, in order to prove Theorem \ref{sgd:thm:minmax_opt}, we only need to show that there exists a mapping $\phi$, an input space $\X$ and a distribution $\rho_\X$ such that the integral operator $K$ introduced in the lemma verifies the assumption on its eigenvalues.
Notice that we show in the proof of Lemma \ref{sgd:prop:step_2} that the universal constant $c_3$ can be taken as $c_3 = 2^{-11}$.

To proceed, let us consider any infinite dimensional Hilbert space ${\cal H}$ with a basis $(e_i)_{i\in\N}$, $\X=\N$ and $\phi:\N\to{\cal H}; i\to \kappa e_i$.
For $a:\N\to\R$ we have
\[
	(Ka)(i) = \sum_{j\in\N} \scap{\phi(i)}{\phi(j)} a(j) \rho(j) = \kappa^2 a(i)\rho(i).
\]
Hence, the eigenvalues of $K$ are $(\kappa^2\rho_\X(i))_{i\leq n}$.
It suffices to consider $\rho_\X(i) = i^{-\alpha} / \zeta(\alpha)$ to conclude.

The eigenvalue decay in $O(i^{-\alpha})$ can also be witnessed in many regression problems.
One way to build those cases is to turn a sequence of non-negative real values into a one-periodic function $h$ from $\R^d$ to $\R$ thanks to the inverse Fourier transform.
Using \cite{Bochner1933}, one can construct a map $\phi$ such that the convolution operator linked with $h$ corresponds to the operator $K$.
When $\rho$ is uniform on $[0,1]^d$, diagonalizing this convolution operator with the Fourier functions and using the property in Lemma \ref{sgd:lem:lin_ell} shows that the class of functions ${\cal F}$ are akin to Sobolev spaces.
Similar behavior can be proven when $\X=\R^d$ and $\rho_\X$ is absolutely continuous with respect to the Lebesgue measure and has bounded density \citep{Widom1963}.
We refer the curious reader to \cite{Scholkopf2001} or \cite{Bach2023} for details.

\section{Unbiased weakly supervised stochastic gradients}
\label{sgd:app:generic}

In this section, we provide a generic scheme to acquire unbiased weakly supervised stochastic gradients, as well as specifications of the formula given in the main text for least-squares and median regression.

\subsection{Generic implementation}

Suppose that $\Theta$ is finite dimensional, or that it can be approximated by a finite dimensional space without too much approximation error.
For example, in the realm of scalar-valued kernel methods, it is usual to consider either the random finite dimensional space $\Span\brace{\phi(x_i)}_{i\leq n}$ for $(x_i)$ the data points, or the finite dimension space linked to the first eigenspaces of the operator $\E[\phi(X)\otimes \phi(X)]$.
In the context of neural networks, the parameter space is always finite-dimensional.

Suppose also that, given $\theta$, we know an upper bound $M_\theta$ on the amplitude of $\nabla_\theta \ell(f_\theta(x), y)$, or that we know how to handle clipped gradients at amplitude $M_\theta$ for SGD.
Then, similarly to the least-squares method proposed in the main text, we can access weakly supervised gradient through the formula
\begin{equation*}
	%	\label{sgd:eq:gen_abs_sgd}
	\nabla_\theta\ell(f_\theta(x), y) = \frac{2M_\theta(\card\Theta^2 + 4\card\Theta + 3)}{\pi^{3/2}} \E_{U \sim \uniform{B_\Theta},V\sim\uniform{[0, M_\theta]}}[\ind{y\in(z\to\scap{U}{\nabla_\theta \ell(f_\theta(x), z)})^{-1}([V, \infty))} U],
\end{equation*}
where $B_\Theta$ is the unit ball of $\Theta$.

This scheme is really generic, and we do not advocate for it in practice as one may hope to leverage specific structure of the loss function and the parametric model in a more efficient way.
This formula is rather a proof of concept to illustrate that our technique can be applied generically, and is not specific to least-squares or median regression.

\subsection{Specific implementations}

Let us prove the two formulas to get stochastic gradients for both least-squares and median regression.
We begin with median regression.
Consider $z\in\mathbb{S}^{m-1}$, and let us denote
\[
	x = \E_U[\sign(\scap{z}{U}) U].
\]
The direction $x/\norm{x}\in\mathbb{S}^{m-1}$ is characterized by the argmax over the sphere of the linear form
\[
	y\to \scap{\E_U[\sign(\scap{z}{U})U]}{y}
	= \E_U[\sign(\scap{z}{U})\scap{U}{y}].
\]
This linear form has a unique maximizer on $\mathbb{S}^{m-1}$ and by invariance by symmetry over the axis $z$, this maximizer is aligned with $z$, hence $x = c_x\cdot z$.
We compute the amplitude with the formula, because $z$ is a unit vector
\[
	c_x = \scap{x}{z} = \E_U[\sign(\scap{z}{U})\scap{U}{z}].
\]
By invariance by rotation of both the uniform distribution and the scalar product, $c_x$ is actually a constant, it is equal to its value $c_2 = c_{e_1}$.

The same type of reasoning applies for the least-squares case.
Consider $z \in\R^m$, and denote
\[
	x = \E_{U, V}\bracket{\ind{\scap{z}{U} \geq V}\cdot U}.
\]
For the same reasons as before $x = c_x\cdot u$ for $u = z / \norm{z}$, and $c_x$ verifies
\begin{align*}
	c_x
	 & = \scap{x}{u}
	= \E_{U,V}[\ind{\scap{z}{U} \geq V}\scap{U}{u}]
	= \E_U[\E_{V}[\ind{\scap{z}{U} \geq V}]\scap{U}{u}]
	\\&= \E_{U}[\ind{\scap{z}{U} > 0} \frac{\scap{z}{U}}{M}\scap{U}{u}]
	= \frac{\norm{z}}{M}\E_{U}[\ind{\scap{u}{U}>0} \scap{U}{u}^2].
\end{align*}
Hence,
\[
	x = \frac{1}{M}\E_{U}[\ind{\scap{u}{U}>0} \scap{U}{u}^2] \cdot z = c_1 \cdot z.
\]
This explains the formula for least-squares.

\begin{lemma}[Constant for the uniform strategy]
	Under the uniform distribution on the sphere
	\begin{equation}
		c_2 = \E_{u\sim\mathbb{S}^{m-1}} \bracket{\abs{\scap{u}{e_1}}}
		= \frac{\sqrt{\pi}\,\Gamma(\frac{m-1}{2})}{m\,\Gamma(\frac{m}{2})} \geq \frac{\sqrt{2 \pi}}{m^{3/2}}.
	\end{equation}
\end{lemma}
\begin{proof}
	Let us compute $c_2 = \E_{u\sim\mathbb{S}^{m-1}} \bracket{\abs{\scap{u}{e_1}}}$.
	This constant can be written explicitly as
	\[
		c_2 = \frac{\int_{x\in\mathbb{S}^{m-1}} \abs{x_1} \diff x}{\int_{x\in\mathbb{S}^{m-1}} \diff x}.
	\]
	Remark that for any function $f:\R\to\R$, we have
	\[
		\int_{\mathbb{S}^{m-1}} f(x_1) \diff x
		= \int_{x_1 \in [-1, 1]} f(x_1)\diff x_1 \int_{\tilde{x}\in \sqrt{1-x_1^2}\cdot\mathbb{S}^{m-2}} \diff \tilde{x}
		= \int_{x_1\in[-1,1]} f(x_1) (1 - x_1^2)^{\frac{m-2}{2}} \diff x_1 \int_{\tilde{x}\in\mathbb{S}^{m-2}} \diff \tilde{x}.
	\]
	By denoting $S_{m}$ the surface of the $m$-sphere, the last integral is nothing but $S_{m-2}$.
	By setting $f(x)=1$, we can retrieve by recurrence the expression of $S_m$.
	In our case, $f(x)=\abs{x}$, so we compute, with $u = 1-x^2$
	\[
		\int_{x_1\in[-1,1]} \abs{x_1} (1 - x_1^2)^\frac{m-2}{2} \diff x_1
		=2\int_{x_1\in[0,1]} x_1 (1 - x_1^2)^\frac{m-2}{2} \diff x_1
		= \int_{u=0}^1 u^\frac{m-2}{2} \diff u = \frac{1}{m}.
	\]
	This leads to
	\[
		c_2 = \frac{S_{m-2}}{m S_{m-1}} = \frac{\sqrt{\pi}\,\Gamma(\frac{m-1}{2})}{m\,\Gamma(\frac{m}{2})}.
	\]
	The ratio $S_{m-2} / S_{m-1}$ can be expressed with the integral corresponding to $f=1$, but it is common knowledge that $S_{m-1} = \sfrac{2\pi^{m/2}}{\Gamma(m/2)}$.
\end{proof}

\begin{lemma}[Constant for least-squares]
	Under the uniform distributions on $[0, M]$ and the sphere
	\begin{equation}
		c_1 = \E_{y\sim[0, M]}\E_{u\sim\mathbb{S}^{m-1}} \bracket{\ind{\scap{u}{e_1} > v} \scap{u}{e_1}} = \frac{\pi^{3/2}}{M (m^2 + 4m + 3)}.
	\end{equation}
\end{lemma}
\begin{proof}
	Similarly to the previous case, this constant can be written explicitly as
	\[
		c_1 = \frac{1}{2}\frac{\int_{y\in[0,M]}\int_{x\in\mathbb{S}^{m-1}} \abs{x_1} \ind{\abs{x_1} > y} \diff y\diff x}{M\int_{x\in\mathbb{S}^{m-1}} \diff x}
		= \frac{\int_{x\in\mathbb{S}^{m-1}} x_1^2 \diff x}{2M\int_{x\in\mathbb{S}^{m-1}} \diff x}.
	\]
	We continue as before with
	\[
		\int_{x_1\in[-1,1]} \abs{x_1}^2 (1 - x_1^2)^\frac{m-2}{2} \diff x_1
		= 2\int_{x\in[0,1]} x^2 (1 - x^2)^\frac{m-2}{2} \diff x
		= \frac{2\pi \Gamma(\frac{m}{2})}{4\Gamma(\frac{m+3}{2})}.
	\]
	This leads to
	\[
		c_1 = \frac{\pi \Gamma(\frac{m}{2})}{4M\Gamma(\frac{m+3}{2})}\cdot\frac{\sqrt{\pi}\,\Gamma(\frac{m-1}{2})}{\Gamma(\frac{m}{2})}
		= \frac{\pi^{3/2} \Gamma(\frac{m-1}{2})}{4M\Gamma(\frac{m+3}{2})}
		= \frac{\pi^{3/2}}{M (m^2 + 4m + 3)}.
	\]
	This is the result stated in the lemma.
\end{proof}
\section{Median surrogate}
\label{sgd:proof:sur}

Let us begin this section by proving Proposition \ref{sgd:prop:sur}.
This result is actually the integration over $x\in\X$ of a pointwise result, so let us fix $x\in\X$.
Consider a probability distribution $p\in\prob{\Y}$ over $\Y$, and its median $\Theta^* \subset \R^\Y$ defined as the minimizer of ${\cal R}_S(\theta) = \E_p[\norm{\theta - e_Y}]$.
We will to prove that $\cup_{\theta\in\Theta^*}\argmax_{y\in\Y} \theta_y = \argmax_{y\in\Y} p(y)$.

Let us begin by the inclusion $\argmax_{y\in\Y} p(y) \subset \cup_{\theta\in\Theta^*}\argmax_{y\in\Y} \theta_y$.
To do so, consider $\theta\in\R^\Y$ and $\sigma \in \Sfrak_\Y$ the transposition of two elements $y$ and $z$ in $\Y$.
Denote by $\theta_\sigma \in \R^\Y$, the vector such that $(\theta_\sigma)_{y'} = \theta_{\sigma(y')}$ for any $y'\in\Y$.
We have
\begin{align*}
	{\cal R}_S(\theta) - {\cal R}_S(\theta_\sigma)
	 & = \sum_{y'\in \Y} p(y') \paren{\norm{\theta - e_{y'}} - \norm{\theta_\sigma - e_{y'}}}
	\\&= \sum_{y'\in \Y} p(y') \paren{\sqrt{\sum_{z'\in\Y}\theta_{z'}^2 + (1 - \theta_{y'})^2 - \theta_{y'}^2} -
		\sqrt{\sum_{z'\in\Y}\theta_{\sigma(z')}^2 + (1 - \theta_{\sigma(y')})^2 - \theta_{\sigma(y')}^2}}
	\\&= (p(y) - p(z)) \paren{\sqrt{\sum_{z'\in\Y}\theta_{z'}^2 + 1 - 2\theta_{y}} -
		\sqrt{\sum_{z'\in\Y}\theta_{z'}^2 + 1 - 2\theta_{z}}}.
\end{align*}
Because, for any $a \in \R_+$, the function $x \to \sqrt{a - 2x}$ is increasing, if $p(y) > p(z)$, then to minimize ${\cal R}$, we should make sure that $\theta_y \geq \theta_z$i.
As a consequence, because of symmetry, the modes of $p$ do correspond to argmax of $(\theta^*_y){y\in\Y}$ for some $\theta^*\in\Theta^*$.

Let us now prove the second inclusion.
To do so, suppose that $p(1) > p(2)$, and let us show that $\theta^*_1 > \theta^*_2$.
Let us parametrize $\theta_1 = a + \epsilon$ and $\theta_2 = a - \epsilon$ for a given $a$, and show that $\epsilon = 0$ is not optimal in order to minimize the risk ${\cal R}_S$ seen as a function of $\epsilon$.
To do so, we can use the Taylor expansion of $\sqrt{1 + x} = 1 + x / 2$.
Hence, with $A = \sum_{y > 2} (\theta^*_y)^2$, retaking the last derivations
\begin{align*}
	{\cal R}_S(\epsilon)
	 & = p(1)\sqrt{(a+\epsilon)^2 + (a-\epsilon)^2 + A + 1 - 2(a + \epsilon)}
	\\&\qquad+ p(2)\sqrt{(a+\epsilon)^2 + (a-\epsilon)^2 + A + 1 - 2(a - \epsilon)}
	\\&\qquad+ \sum_{y>2} p(y)\sqrt{(a+\epsilon)^2 + (a-\epsilon)^2 + A + 1 - 2\theta^*_y}
	\\&= p(1)\sqrt{2a^2+ 2\epsilon^2 + A + 1 - 2a - 2\epsilon}
	\\&\qquad+ p(2)\sqrt{2a^2+2\epsilon^2 + A + 1 - 2a + 2\epsilon} + c + o(\epsilon)
	\\&= \tilde{c} + \frac{\epsilon}{\sqrt{2a^2 + A + 1 - 2a}}(p(2) - p(1)) + o(\epsilon).
\end{align*}
This shows that taking $\theta_1^* = \theta_2^*$, that is $\epsilon=0$, is not optimal, hence we have the second inclusion, which ends the proof.
Note that we have proven a much stronger result, we have shown that $(\theta_y)$ and $p(y)$ are order in the exact same fashion (with respect to the strict comparison $p(y)>p(z)\Rightarrow \theta_y^* > \theta_z^*$ for any $\theta^*\in\Theta^*$).

\subsection{Discussion around the median surrogate.}
The median surrogate have some nice properties for a surrogate method, in particular it does not fully characterize the distribution $p(y)$ in the sense that there is no one-to-one mapping from $p$ to $\theta^*$.
For example, when $\Y = \brace{1, 2, 3}$ if
\(
p(y=e_1), p(y=e_2), p(y=e_3) \propto (1, 1, 2\cos(\pi / 6)),
\)
then the geometric median correspond to $\theta^* = e_3$.
This differs from smooth surrogates, such as logistic regression or least-squares, that implicitly learn the full distribution $p$, which should be seen as a waste of resources.
Non-smooth surrogates tend to exhibit faster rates of convergence (in terms of decrease of the original risk as a function of the number of samples) than smooth surrogates when rates are derived through calibration inequalities \citep{Nowak2021}.
It would be nice to derive generic calibration inequality for the median surrogate for multiclass, and see how to derive a median surrogate for more structured problems such as ranking problems.

\begin{figure}[ht]
	\centering
	\includegraphics{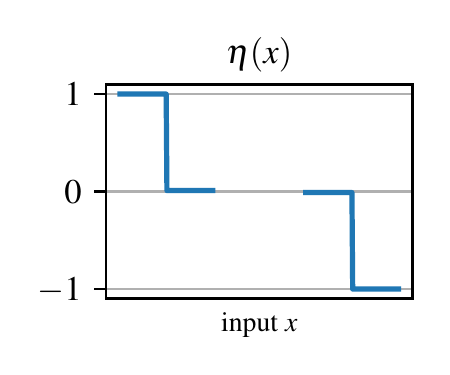}
	\includegraphics{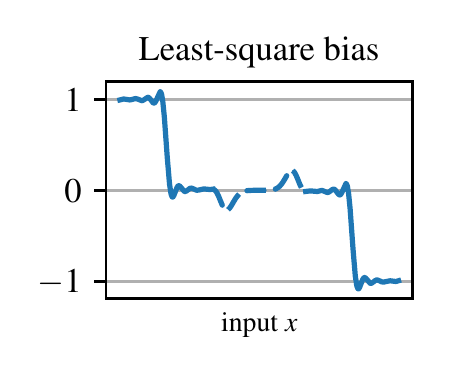}
	\includegraphics{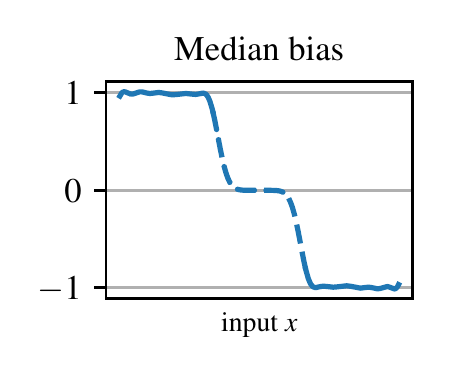}
	\includegraphics{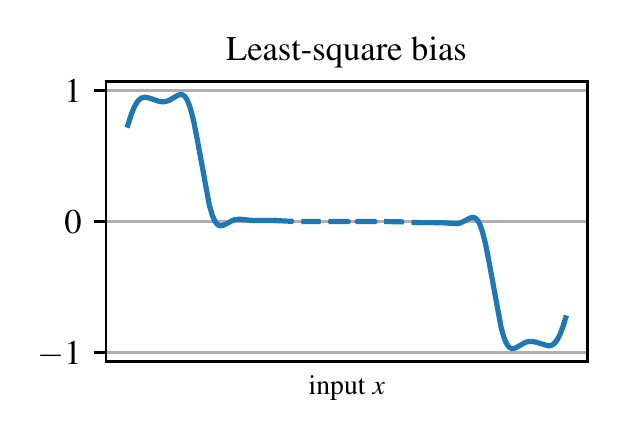}
	\includegraphics{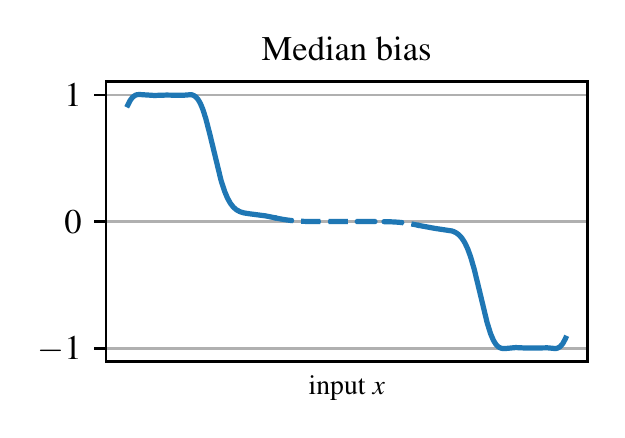}
	\caption{
	{\em Comparison of least-squares and absolute deviation with noise irregularity} for a classification problem specified by $\X = [0, 3]$, $\Y = \brace{-1, 1}$ with $X$ uniform on $[0, 1]\cup[2, 3]$ and $\eta(x) = \E\brace{Y\midvert X=x}$ specified on the left figure.
	The optimal classifier, with respect to the zero-one loss, $f^*(x) = \sign \eta$ takes value one on $[0, 1]$ and value minus one on $[2, 3]$.
	The regularized solution are defined as $\argmin_g \E[\norm{\scap{\phi(X)}{\theta} - Y}^p] + \lambda \norm{\theta}$ with $p=2$ for least-squares (middle), and $p=1$ for the median (right).
	They can be translated into classifiers with the decoding $f = \sign g$.
	In this figure, we choose $\phi$ implicitly through the Gaussian kernel $k(x, x') = \scap{\phi(x)}{\phi(x')} = \exp(-\norm{x-x'}^2 / 2\sigma^2)$ with $\sigma = .1$ which explains the frequency of the observed oscillations, and choose $\lambda = 10^{-6}$ (top) and $\lambda = 10^{-2}$ (bottom).
	On the one hand, because the least-squares surrogate is trying to estimate $\eta$ it suffers from its lack of regularity, leading to Gibbs phenomena that restricts it to be a perfect classifier.
	On the other hand, the absolute deviation is trying to approach the function $f^*$ itself, and does not suffer from its lack of regularity.
	In this setting, if we approach the original classification problem by minimization of the surrogate empirical risks, and denote by $g_n$ this minimizer and $f_n =\sign g_n$ its decoding, $f_n$ obtained through median regression will converge exponentially fast toward $f^*$, while $f_n$ obtained through least-squares will never converge to the solution $f^*$.
	}
	\label{sgd:fig:med_ls}
\end{figure}

\begin{figure}[ht]
	\centering
	\includegraphics{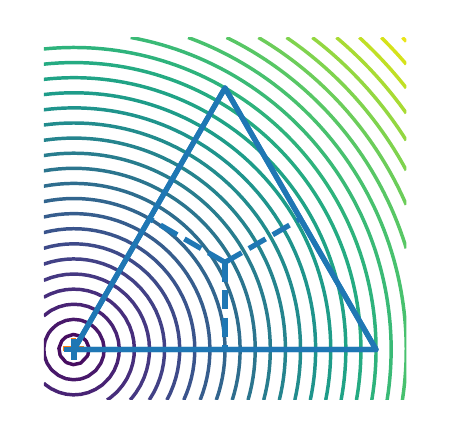}
	\includegraphics{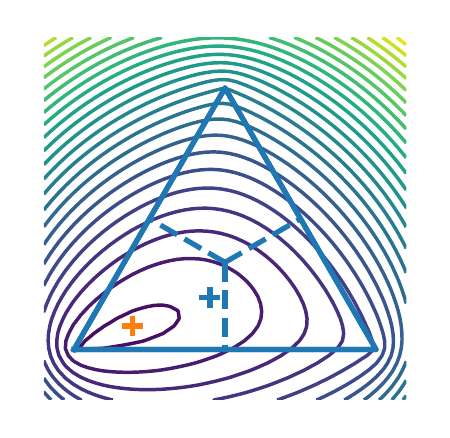}
	\includegraphics{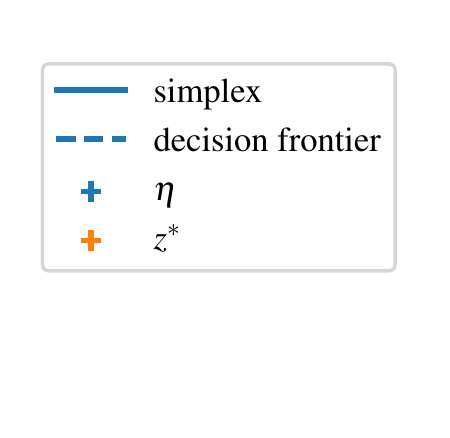}
	\caption{
	{\em Comparison of least-squares and median surrogate without context.}
	Consider a context-free classification problem that consists in estimating the mode of a distribution $p\in\prob{\Y}$, or equivalently the minimizer of the 0-1 loss.
	Such a problem can be visualized on the simplex $\prob{m}$ where $\Y = \brace{y_1, \cdots, y_m} \simeq \brace{1, \cdots, m}$ is mapped to the canonical basis $\brace{e_i}_{i\in[m]} \in \R^m$.
	The figure illustrates the case $m=3$.
	The least-squares and median surrogate methods can be understood as working in this simplex, estimating a quantity $z \in \prob{\Y}$, before performing the decoding $y(z) = \argmax_y \scap{z}{e_y}$.
	Such a decoding partitions the simplex in regions whose frontiers are represented in dashed blue on the figure.
	The distribution $p$ is characterized on the simplex by $\eta = \E_{Y\sim p}[e_Y] = \argmin \E_{Y\sim p} [\norm{z - e_Y}^2]$.
	This quantity $\eta$ is exactly the quantity estimated by the least-squares surrogate.
	The median surrogate searches the minimizer $z^*$ of the quantity ${\cal E}(z) = \E_{Y\sim p}[\norm{z - e_Y}]$, whose level lines are represented in solid on the figure.
	One of the main advantage of the median surrogate compared to the least-squares one is that $z^*$ is always farther away from the boundary frontier than $\eta$, meaning that for a similar estimation error on this quantity, the error on the decoding, which corresponds to an estimate of the mode of $p$, will be much smaller for the median surrogate.
	The left figure represents the case $p = (1, 0, 0)$, the right figure the case $p =(.45, .35, .2)$.}
	\label{sgd:fig:med_simplex}
\end{figure}

\begin{figure}[ht]
	\centering
	\includegraphics{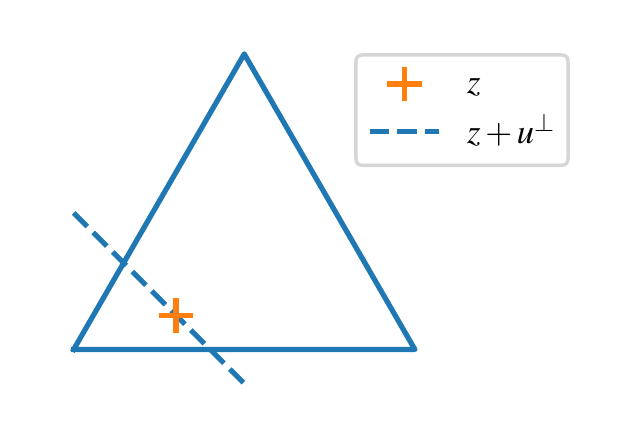}
	\caption{
		{\em Query strategy based on regression surrogate.}
		Retaking the simplex representation of Figure \ref{sgd:fig:med_simplex}, the query strategy for classification approached with least-squares surrogate or median surrogate consists in looking at the current surrogate estimate $z$ in the simplex $\prob{\Y}$, taking a random direction $u\in\R^\Y$ and querying $\sign(\scap{e_Y - z}{u})$.
		We see that with three elements, when $Y$ is deterministic, the optimal query strategy consists in considering $s =\brace{y}$, while surrogate strategies, such as least-squares and median regression, that learn $z^* = e_y$, would only make such a query only two third of the time (which is the ratio of the solid angle of $[e_2, e_3]$ from $e_1$ divided by $\pi$).
		This shows that those surrogate strategies do not fully leverage the specific structure of the output.
	}
	\label{sgd:fig:med_query}
\end{figure}

\section{Classification with a min-max game}
\label{sgd:proof:minmax}

In this section, we prove and extend on Proposition \ref{sgd:prop:minmax}.
First of all, let us consider the average loss, for $(v_y)\in\R^\Y$ summing to one
\[
	\bar L(v, s) = 1 - \sum_{y\in s} v_y = \sum_{y\notin s} v_y.
\]
Consider now this loss conditioned on the observation $\ind{y\in s}$, we have plenty of characterizations of $L$,
\begin{align*}
	L(v, s; \ind{y\in s} - \ind{y\notin s})
	 & = \ind{y\in s} \bar L(v, s) + \ind{y\notin s}\bar L(v, \Y\setminus s)
	= \ind{y\in s}\sum_{y\notin s} v_y + \ind{y\notin s}\sum_{y\in S} v_y
	\\&= \ind{y\in s} + (\ind{y\notin s} - \ind{y\in s})\sum_{y\in s} v_y
	= \ind{y\notin s} + (\ind{y\in s} - \ind{y\notin s})\sum_{y\notin s} v_y
	\\&= \frac{1}{2} - \frac{1}{2} (\ind{y\in s} - \ind{y\notin s})\paren{\sum_{y\in s} v_y - \sum_{y\notin s} v_y}
	= \frac{1}{2} + \frac{1}{2} (\ind{y\in s} - \ind{y\notin s})\paren{1 - 2\sum_{y\in s} v_y}.
\end{align*}
Minimizing this loss or the loss $2L-1$ as defined in Proposition \ref{sgd:prop:minmax} is equivalent.

\subsection{Consistency}
Let us consider the loss as defined in this proposition, we have the characterization
\[
	L(v, s; \ind{y\in s} - \ind{y\notin s}) = (\ind{y\in s} - \ind{y\notin s})\paren{\sum_{y\in s} v_y - \sum_{y\notin s} v_y}.
\]
Let us rewrite~\eqref{sgd:eq:minmax} based on this previous characterization of the loss, we have
\[
	\E_Y[L(v, s, \ind{Y\in s} - \ind{Y\notin s})]
	= -(\Pbb_Y(Y\in s) - \Pbb_Y(Y\notin s))\paren{\sum_{y\in s} v_y - \sum_{y\notin s} v_y}.
\]
Hence, without any context variable, the min-max game \eqref{sgd:eq:minmax} can be rewritten as
\begin{equation}
	\min_{v\in\prob{\Y}} \max_{\mu\in\prob{\cal S}}
	- \sum_{s\in{\cal S}} \mu_s (\Pbb_Y(Y\in s) - \Pbb_Y(Y\notin s))\paren{\sum_{y\in s} v_y - \sum_{y\notin s} v_y}.
\end{equation}
We will analyze this problem through the lens of a mix-actions zero-sum game.
We know from \cite{VonNeumann1944} that a solution to this min-max problem exists, and that one can switch the min-max to a max-min without modifying the value of the solution.
Let us denote by $(v^*,\mu^*)$ the argument of a solution.
To minimize the value of this game, the player $v$ should play such that
\[
	\sign(\sum_{y\in s} v_y^* - \sum_{y\notin s} v_y^*)
	= \sign (\Pbb(Y\in s) - \Pbb(Y\notin s))
	= \sign (\sum_{y\in s} \Pbb(Y=y) - \sum_{y\notin s} \Pbb(Y=y)),
\]
which allows this player to ensure a negative value to the game.
Stated otherwise
\begin{equation}
	\label{sgd:eq:loss_imp}
	\forall\, s\in{\cal S}, \qquad
	\Pbb(Y\in s) > \frac{1}{2}\quad\Rightarrow\quad\sum_{y\in s} v_y^* \geq \frac{1}{2}.
\end{equation}
As a consequence, if there exists any set such that $\Pbb(Y\in s) = 1/2$, the best strategy of player $\mu$ is to play only those sets to ensure the value zero, and any $v$ that satisfies~\eqref{sgd:eq:loss_imp} is optimal.
It should be noted that~\eqref{sgd:eq:loss_imp} does not generally imply that $(v_y)_{y\in\Y}$ has the same ordering as $(\Pbb(Y=y))_{y\in\Y}$.

When $\brace{y^*}\in{\cal S}$ and $\Pbb(Y=y^*) > 1/2$, if $v = \delta_{y^*}$, the prediction player is able to ensure a value of $\max_{s\in{\cal S}}-\abs{2\Pbb(Y\in s) - 1}$, which is maximized by the query player with $s = \brace{y^*} \cup s'$ for any $s'$ such that $\Pbb(Y\in s') = 0$. Other strategies for $v$ will only increase this value, hence $v^* = \delta_{y^*}$ which implies the first part of Proposition \ref{sgd:prop:minmax}.

\paragraph{A counter example.}
While we hope that the solution $(v^*, \mu^*)$ does characterize the original solution $y^*$, it should be noted that $v^*$ alone does not characterize $y^*$.
Indeed, it is even possible to have $v^*$ uniquely defined without having $y^* = \argmax_{y\in\Y} v^*_y$.
For example, consider the case where $\Y = \brace{1, 2, 3}$ and $(\Pbb(Y=i))_{i\in[3]} = (.4, .3, .3)$.
By symmetry, the player $\mu$ only has to play on ${\cal S} = \brace{\brace{1}, \brace{2}, \brace{3}}$, which leads to the min-max game
\[
	\min_{v} \max_{\mu}
	\paren{
		\begin{array}{c}
			\mu_{\brace 1} \\
			\mu_{\brace 2} \\
			\mu_{\brace 3} \\
		\end{array}
	}^\top
	\paren{
		\begin{array}{ccc}
			.2  & -.2 & -.2 \\
			-.4 & .4  & -.4 \\
			-.4 & -.4 & .4  \\
		\end{array}
	}
	\paren{
		\begin{array}{c}
			v_1 \\
			v_2 \\
			v_3 \\
		\end{array}
	}.
\]
The value of this game is $-.1$ and is achieved for $\mu^* = (.5, .25, .25)$, $v^* = (.25, .375, .375)$.

\subsection{Optimization procedure}

Let us rewrite the problem through the objective
\[
	{\cal E}(g, \mu) = \E_{(X, y)\sim \rho}\E_{S\sim\mu(x)}[L(g(X), S, \ind{Y\in S} - \ind{Y\notin S})].
\]
We want to solve the min-max problem $\min_g\max_\mu{\cal E}(g, \mu)$.
This problem can be solved efficiently based on the vector field point of view of gradient descent \citep{Bubeck2015} if:
\begin{itemize}
	\item we can parametrize the function $g:\X\to\prob\Y$ such that ${\cal E}$ is convex with respect to the parametrization of $g$;
	\item we can access unbiased stochastic gradients of ${\cal E}$ with respect to $g$ that have a small second moment;
	\item we can parametrize the function $\mu:\X\to\prob{\cal S}$ such that ${\cal E}$ is concave with respect to the parametrization of $\mu$;
	\item we can access unbiased stochastic gradients of ${\cal E}$ with respect to $\mu$ that have a small second moment.
\end{itemize}
The first two points are no problems, $g$ can be parametrized with softmax regression, and since $L$ is linear with respect to the scores, it will keep the problem convex.
Moreover, to access a stochastic gradient of ${\cal E}$, one can sample $X_i\sim\rho_\X$ and $S_i\sim\mu(X_i)$ before querying $\ind{Y_i\in S_i}$ and computing the gradient of $L(g(X_i), S_i, \ind{Y_i\in S_i} - \ind{Y_i\notin S_i})$ with respect to the parametrization of $g$.

The third point is slightly harder to tackle.
Since ${\cal E}$ is linear with respect to $\mu$, one way to proceed is to find a linear parametrization of $\mu$.
In particular, one can take a family $(g_i)_{i\in[N]}$ of linearly independent functions from $\X$ to $\prob{\cal S}$ and search for $g$ under the form $\sum_{i\in[N]} c_i g_i$ for $(c_i)$ positive summing to one.
To build such a family, one can eventually use ``atom functions'' and simple operations such as symmetry with respect to $\Y$ and ${\cal S}$, rescaling, translation, rotations with respect to $\X$.
For example if $\X$ is a Banach space, one could define atom functions as, for $y_i\in\Y$
\[
	g_i: x\to \frac{\norm{x}}{1+\norm{x}} \frac{1}{\card{\cal S}} \sum_{s\in{\cal S}} e_s + \frac{1}{1+\norm{x}} e_{\brace{y_i}}.
\]
Those functions could be rescaled and translated as $g_{\sigma, \tau, i}(x) = g_i(\sigma (x-\tau))$, in order to specify a family $(g_{\sigma, \tau, i})$ from few values for $\tau$ and $\sigma$.

The last point is the most difficult one.
Without context variables, and with no-parametrization for $\mu$, a naive unbiased gradient strategy for $\mu$ consists in asking random questions to update the full knowledge of $(\Pbb(Y\in s))_{s\in{\cal S}}$.
But such a strategy will be much worse than our median surrogate technique with queries $\ind{Y\in\brace{y}}$ for $y$ sampled uniformly at random in $\Y$.
Eventually, one should go for a biased gradient strategy, while making sure to update $\mu$ coherently to avoid getting stalled on bad estimates as a result of biases.
\begin{figure}[ht]
	\centering
	\includegraphics[width=.7\textwidth]{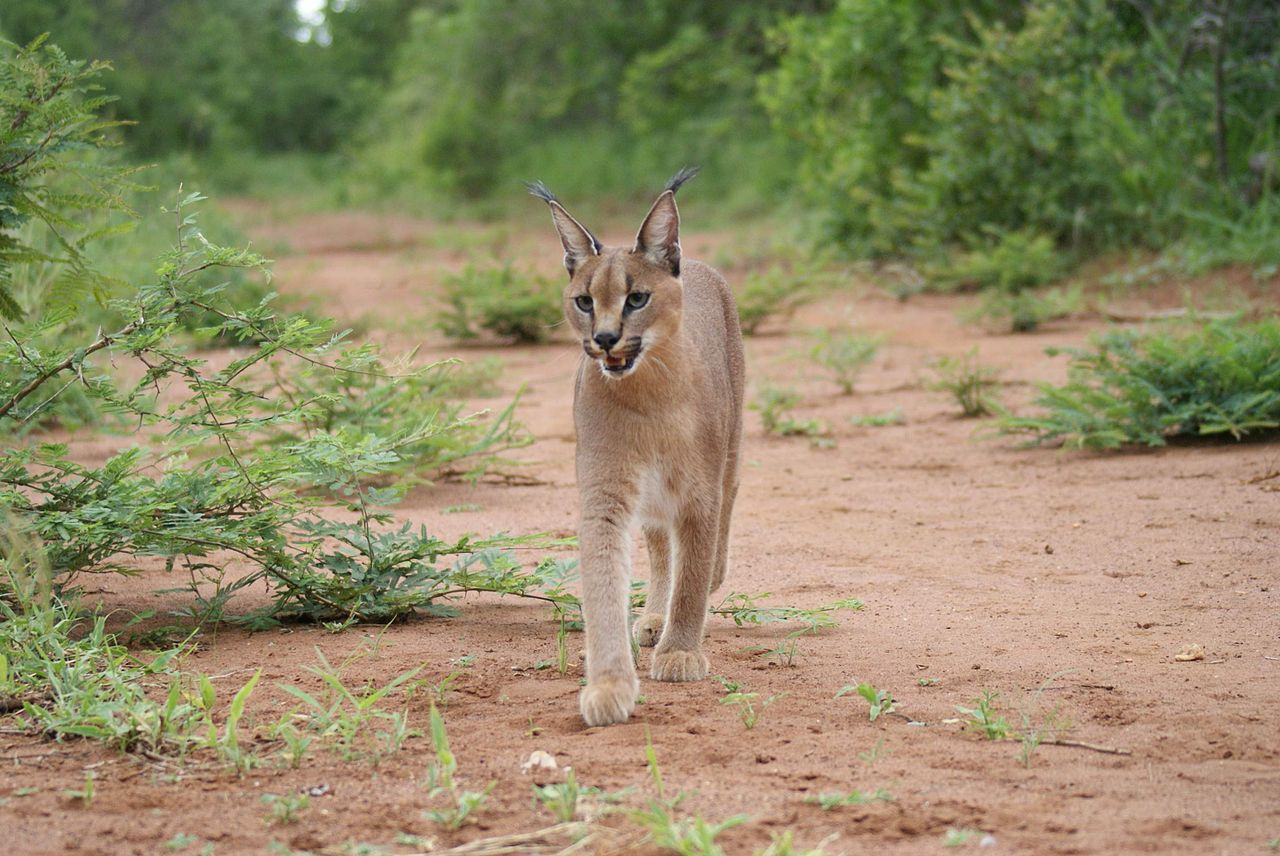}
	\caption{Recognizing fine-grained classes is difficult, but recognizing attributes is easy.}
	\label{sgd:fig:caracal}
\end{figure}
  
\section{Experimental details}
\label{sgd:app:experiments}

Our experiments are done in {\em Python}. We leverage the {\em C} implementation of high-level array instructions by \cite{Harris2020}, as well as the visualization library of \cite{Hunter2007}.
Randomness in experiments is controlled by choosing explicitly the seed of a pseudo-random number generator.

\begin{figure}[ht!]
	\centering
	\includegraphics{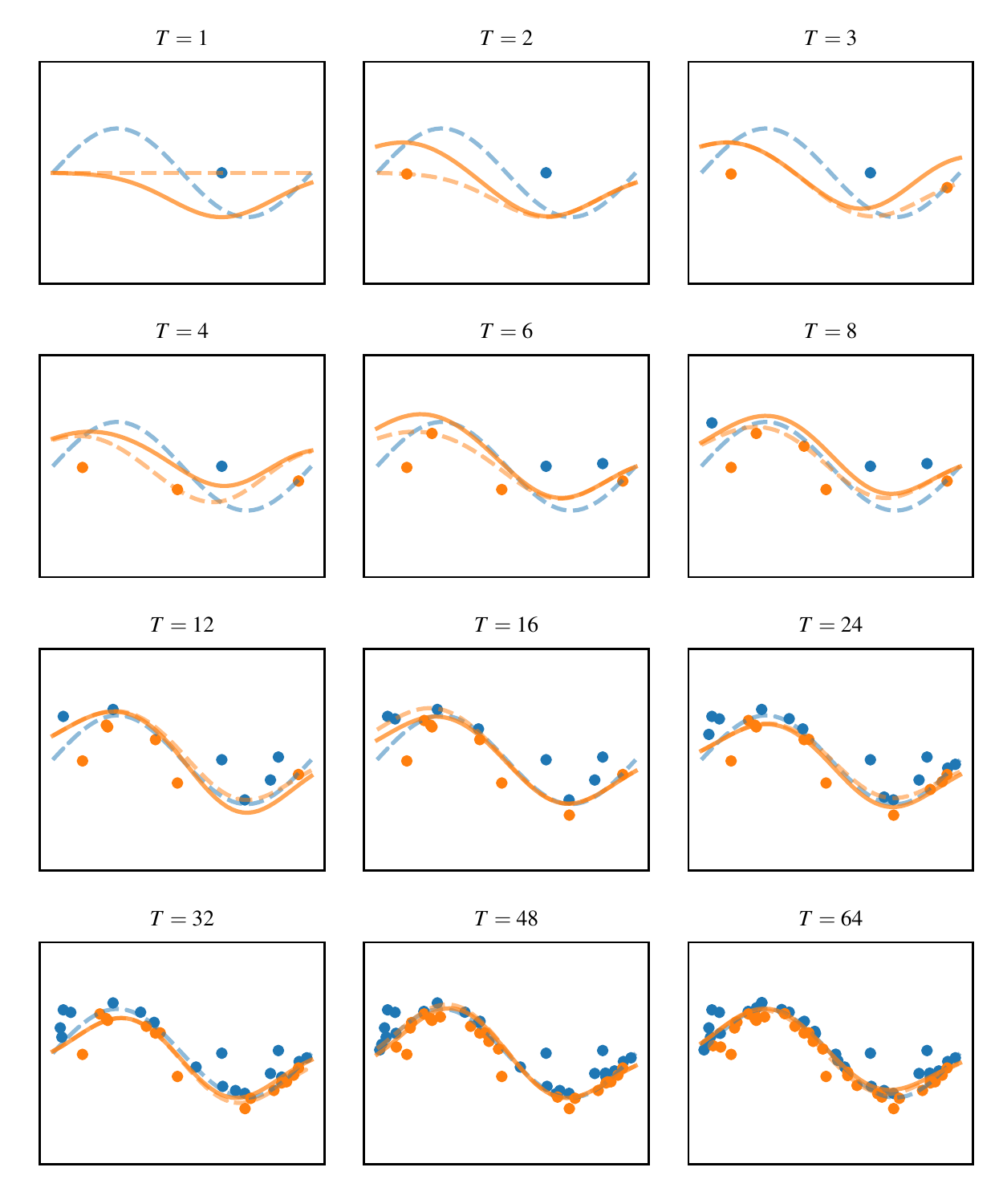}
	\caption{
		{\em Streaming history of the active strategy} to reconstruct the signal in dashed blue in the same setting as Figure \ref{sgd:fig:exp_1}.
		At any time $t$, a point $X_t$ is given to us, our current estimate of $\theta_t$ plotted in dashed orange gives us $z = f_{\theta_t}(X_t)$, and we query $\sign(Y_t - z)$.
		Based on the answer to this query, we update $\theta_t$ to $\theta_{t+1}$ leading to the new estimate of the signal in solid orange.
		In this figure, we see that it might be useful for the practitioners in a streaming setting to reduce the bandwidth of $\phi$ as they advance in time.}
	\label{sgd:fig:exp_1_app}
\end{figure}

\subsection{Comparison with fully supervised SGD}

\begin{figure}[t]
	\centering
	\includegraphics{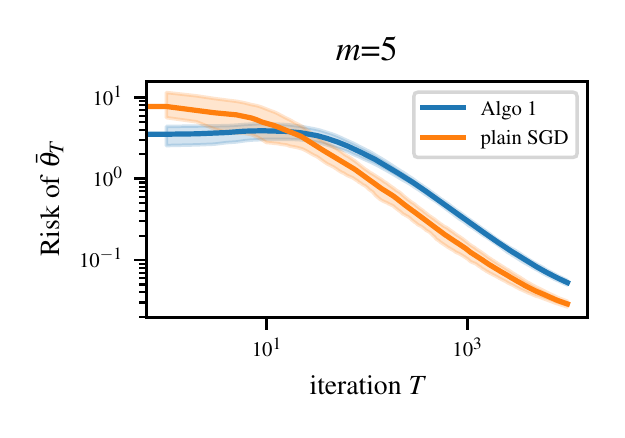}
	\includegraphics{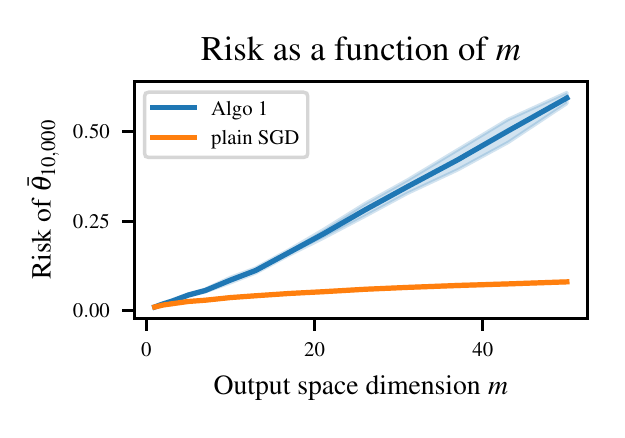}
	\caption{
	{\em Comparison of generalization errors of weakly and fully supervised SGD} as a function of the annotation budget $T$ and output space dimension $m$.
	The setting is similar to Figure \ref{sgd:fig:exp_2}.
	We observe a transitory regime before convergence rates follows the behavior described by Theorem \ref{sgd:thm:sgd}.
	The right side plots the error of both procedures after 10,000 iterations as a function of the output space dimension $m$ between 1 and 50.
	The number of iteration ensures that, for all values of $m\in [50]$, the reported error is well characterized by our theory, in other terms that we have entered the regime described by Theorem \ref{sgd:thm:sgd}.}
	\label{sgd:fig:plain_sgd}
\end{figure}

In this section, we investigate the difference between weakly and fully supervised SGD.
According to Theorem \ref{sgd:thm:sgd}, we only lost a constant factor of order $m^{3/2}$ in our rates compared to fully supervised (or plain) SGD.
This behavior can be checked by adding the plain SGD curve on Figure \ref{sgd:fig:exp_2}. 
On the left side of Figure \ref{sgd:fig:plain_sgd}, we do observe that the risk of both Algorithm \ref{sgd:alg:sgd} and plain SGD decrease with same exponent with respect to number of iteration but with a different constant in front of the rates: that is we observe the same slopes on the logarithm scaled plot, but different intercepts.
Going one step further to check the tightness of our bound, one can plot the intercept, or the error achieved by both Algorithm \ref{sgd:alg:sgd} and plain SGD as a function of the output space dimension $m$. 
The right side of Figure \ref{sgd:fig:plain_sgd} shows evidence that this error grows as $m^\epsilon$ for some $\epsilon \in [1, 3/2]$, which is coherent with our upper bound.
Similarly to Figure \ref{sgd:fig:exp_2}, this figure was computed after cross validation to find the best scaling of the step sizes for each dimension $m$.

\subsection{Passive strategies for classification}
A simple passive strategy for classification based on median surrogate consists in using the active strategy with coordinates sampling, that is $u$ being uniform on $\brace{e_y}_{y\in\Y}$, where $(e_y)_{y\in\Y}$ is the canonical basis of $\R^\Y$ used to define the simplex $\prob{\Y}$ as the convex hull of this basis.
Querying $\ind{\scap{g_\theta(x) - e_y}{e_y} > 0}$ is formally equivalent to the query of $\ind{Y = y}$ when $g_\theta(x) \in \prob{\Y}$.
This is the baseline we plot on Figure \ref{sgd:fig:exp_2}.

\begin{figure}[ht]
	\centering
	\includegraphics{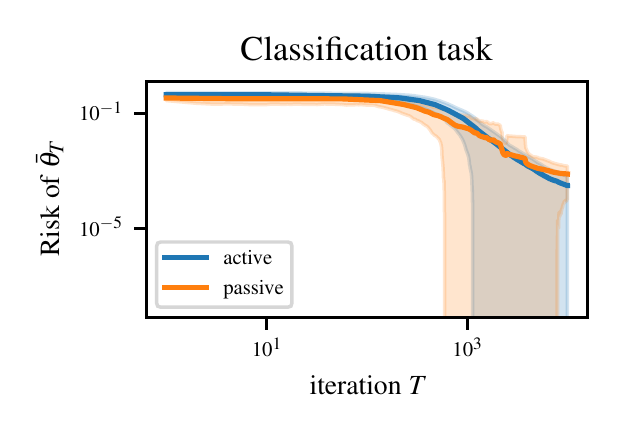}
	\caption{
		{\em Comparison with the infimum loss with better conditioned passive supervision} in a similar setting to Figure \ref{sgd:fig:exp_2} yet with $m=10$, $\epsilon=0$, that is $X$ uniform on $\X$, and $\gamma_0 = 7.5$ for the active strategy and $\gamma_0 = 15$ for the passive strategy.
		We see no major differences between the active strategy based on the median surrogate and the passive strategy based on the median surrogate with the infimum loss.
		Note that the standard deviation is sometimes bigger than the average of the excess of risk, explaining the dive of the dark area on this logarithmic-scaled plot.
	}
	\label{sgd:fig:exp_2_app}
\end{figure}

A more advanced passive baseline is provided by the infimum loss \citep{Cour2011,Cabannes2020}.
It consists in solving
\[
	\argmin_{f:\X\to\Y}{\cal R}_I(f):=\E_{(X, Y)\sim\rho}\E_{S}\bracket{L(f(X), S, \ind{Y\in S})},
\]
where $S$ is a random subset of $\Y$ and $L$ is defined from the original loss $\ell:\Y\times\Y\to\R$ as, for $z\in\Y$, $s\subset \Y$ and $y\in\Y$,
\[
	L(z, s, \ind{y\in s}) = \left\{\begin{array}{cl} \inf_{y'\in s} \ell(z, y') & \text{if } y\in s \\ \inf_{y'\notin s} \ell(z, y') & \text{otherwise.}\end{array}\right.
\]
Random subsets $S$ could be generated by making sure that the variable $(y \in S)_{y\in\Y}$ are independent balanced Bernoulli variables; and by removing the trivial sets $S = \emptyset$ and $S = \Y$ from the subsequent distribution.
In order to optimize this risk in practice, one can use a parametric model and a surrogate differentiable loss together with stochastic gradient descent on the empirical risk.
For classification with the 0-1 loss, we can reuse the surrogate introduced in Proposition \ref{sgd:prop:sur} and minimize, assuming that we always observed $\ind{Y_i\in S_i} = 1$ for simplicity,
\[
	\hat{\cal R}_{I, S}(\theta) = \sum_{i=1}^n \inf_{y \in S_i} \norm{g_\theta(X_i) - e_y}.
\]
Stochastic gradients are then given by, assuming ties have no probability to happen,
\[
	\nabla_\theta \inf_{y \in S_t} \norm{g_\theta(X_t) - e_y}
	= \paren{\frac{g_\theta(X_t) - e_{y^*}}{\norm{g_\theta(X_t) - e_{y^*}}}}^\top D g_\theta(X_t)
	\quad\text{with}\quad y^* := \argmax_{y\in S_t} \scap{g_\theta(X_t)}{e_y}.
\]
This gives a good passive baseline to compare our active strategy with.
In our experiments with the Gaussian kernel, see Figure \ref{sgd:fig:exp_2_app} for an example, we witness that this baseline is highly competitive.
Although we find that it is slightly harder to properly tune the step size for SGD, and that the need to compute an argmax for each gradient slows-down the computations.

\subsection{Real-world classification datasets}

\begin{figure}[ht]
	\centering
	\includegraphics{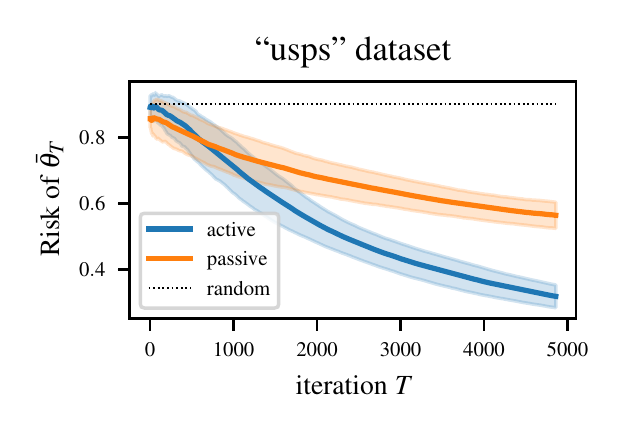}
	\includegraphics{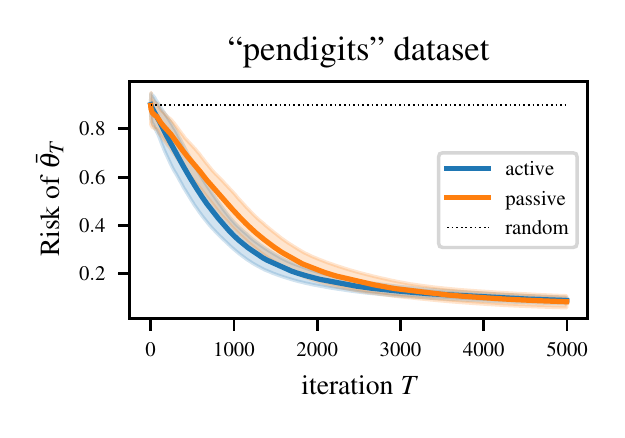}
	\caption{
		{\em Testing errors on two LIBSVM datasets} with a similar setting to Figure \ref{sgd:fig:exp_2_app}.
		Those empirical errors are reported after averaging over 100 different splits of the datasets.
		The step size parameter was optimized visually, which led to $\gamma_0=15$ for the active strategy on ``USPS'', $\gamma_0=60$ for the passive one, $\gamma_0=7.5$ for the active strategy on ``pen digits'', $\gamma_0=30$ for the passive one.
		The dotted line represents ${\cal R} = 1 - m^{-1}$ which is the performance of a random model.
	}
	\label{sgd:fig:exp_libsvm}
\end{figure}

In Figure \ref{sgd:fig:exp_libsvm}, we compare the ``well-conditioned'' passive baseline with our active strategy on the real-world problems of LIBSVM \citep{Chang2011}.
We choose the ``USPS'' and ``pen digits'' datasets as they contain $m=10$ classes each with $n=7291$ and $n=7494$ samples respectively, with $d=50$ and $d=16$ features each.
We have chosen those datasets as they present enough classes that leads to many different sets $S$ to query, and they are made of the right number of samples to do some experiments on a laptop without the need for ``advanced'' computational techniques such as caching or low-rank approximation \citep{Meanti2020}.
On Figure \ref{sgd:fig:exp_libsvm}, we use the same linear model as for Figure \ref{sgd:fig:exp_2}, that is a Gaussian kernel.
We choose the bandwidth to be $\sigma = d/5$, and we normalize the features beforehand to make sure that they are all centered with unit variance.
We report error by taking two thirds of the samples for training and one third for testing, and averaging over one hundred different ways of splitting the datasets.
We observe that the active strategy leads to important gains on the ``USPS'' dataset, yet is not that useful for the ``pen digits'' dataset.
We have not dug in to understand those two different behaviors.

\subsection{Real-world regression dataset \& Nystr\"om method}

In this section, we provide two experiments on real-world datasets.

In order to deal with big regression datasets, it is useful to approximate the parameter space $\Y\otimes{\cal H}$ in Assumption \ref{sgd:ass:source} with a small dimensional space.
To do so, let us remark that given samples $(X_i)_{i\leq n} \in \X^n$ for $n\in\N$, we know that our estimate $f_{\theta_n}$ can be represented as
\[
	f_{\theta_n}(\cdot) = \sum_{i\leq n}\sum_{j\leq m} a_{ij} \scap{\phi(x_i)}{\phi(\cdot)} e_j,
\]
for some $(a_{ij}) \in \R^{p\times m}$ and where $(e_j)_{j\leq m}$ is the canonical basis of $\Y = \R^m$.
For large datasets, that is when $n$ is large, it is smart to approximate this representation through the parameterization
\[
	f_{a}(x) = \sum_{i\leq p}\sum_{j\leq m}a_{ij} k(x, x_i) e_j,
\]
where $p\leq n$ is the rank of our approximation, and $k$ is the kernel defined as $k(x, x') = \scap{\phi(x)}{\phi(x')}$.
Stated with words, we only use a small number $p$, instead of $n$, of vectors $\phi(x_i)$ to parameterize $f$.
This allows to only keep a matrix of size $p\times m$ in memory instead of $n\times m$, while not fundamentally changing the statistical guarantee of the method \citep{Rudi2015}.
In this setting, the stochastic gradients are specified from the fact that 
\[
	u^\top D_a f_a(x) = (u_j k(x, x_i))_{i,j} \in \R^{p\times m}.
\]
In other terms, in order to update the parameter $a$ with respect to the observation made at $(x, u)$, we check how much each coordinate of $a$ determines the value of $u^\top f_a(x)$.

In the following, we experiment with two real-world datasets.
In order to learn the relation between inputs and outputs, we use a Gaussian kernel after normalizing input features so that each of them has zero mean and unit variance.
To keep computational cost, we sample $p$ random (Nystr\"om) representers among the training inputs which are used to parameterize functions.
To avoid overfitting, we add a small regularization to the empirical objective. It reads $\lambda \norm{\theta}_{{\cal H}}^2$ with our notations and corresponds to the Hilbertian norm inherited from the reproducing kernel $k$ of the function $f_{\theta}$ \citep{Scholkopf2001}.

\begin{figure}[h]
	\centering
	\includegraphics{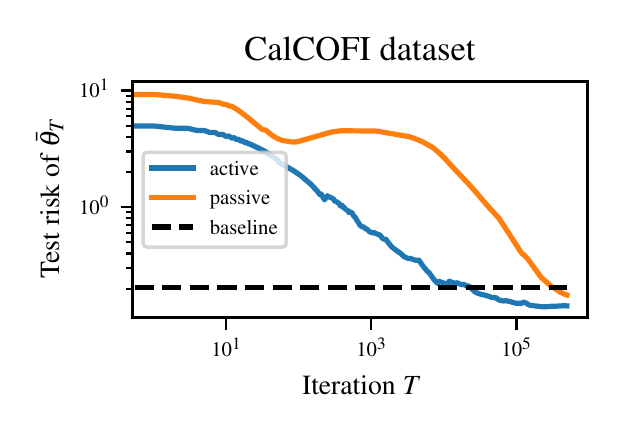}
	\includegraphics{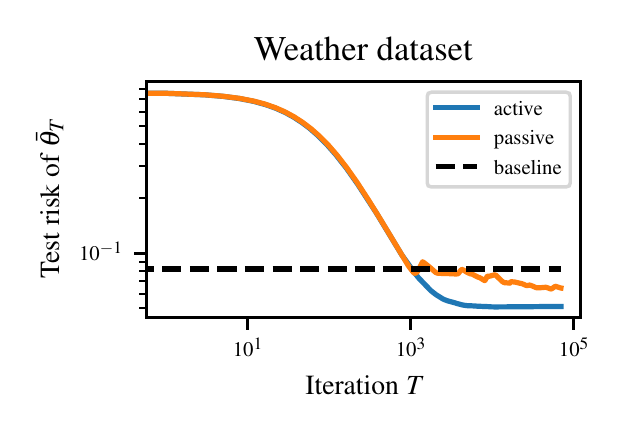}
	\caption{
	{\em Testing error on two real-world regression datasets.}
	On both datasets, a single pass was made through the data in a chronological fashion, and errors were computed from the 26,453 most recent data samples for the ``Weather'' dataset, and from a random sample of 10,000 samples among the 155,140 most recent samples for the ``CalCOFI'' dataset.}
	\label{sgd:fig:real_world}
\end{figure}

Our first experiment is based on the data collected by the California Cooperative Oceanic Fisheries Investigation between March 1949 and November 2016.\footnote{CalCOFI data is licensed under the CC BY 4.0 license and the data is available at \url{https://calcofi.org/}.}
It consists of more than 800,000 seawater samples including measurements of nutriments (set aside in our experiments) together with pressure, temperature, salinity, water density, dynamic height (providing five input parameters), as well as dissolved oxygen, and oxygen saturation (the two outputs we would like to predict).
We assume that we can measure if any weighted sum of oxygen concentration and saturation is above a threshold by letting some population of bacteria evolves in the water sample and checking if it survives after a day.
If the measurements are done on the day of the sample collection, this setting exactly fits in the streaming active labeling framework.
After cleaning the dataset for missing values, the dataset contains 655,140 samples.
The ``CalCOFI'' dataset results are reported on the left of Figure \ref{sgd:fig:real_world}, parameters were chosen as $p=100$, $\sigma = 10$, $\lambda = 10^{-6}$ and $\gamma_0 = 1$.
For the passive strategy, random queries were chosen to follow a normal distribution with the same mean as the targets and one third of their standard deviation ({\em i.e.} we ask if the apparent temperature is lower than the usual one plus or minus a perturbation).
The plotted baseline corresponds to linear regression performed over the entire dataset.
It takes about 10,000 samples for our active strategy to be competitive with this baseline, and 200,000 samples for the passive one.

The second experiment makes use of data collected through the Dark Sky API (which is now part of Apple WeatherKit). 
It is made of 96,454 weather summaries between 2006 and 2016 in the city of Szeged, Hungary.
Our task consists in computing the apparent temperature from real temperature, humidity, wind speed, wind bearing, visibility and pressure.
The apparent temperature is an index that searches to quantify the subjective feeling of heat that humans perceive, it is expressed on the same scale as real temperature.
One way to measure it would be to ask some humans if the outside is hotter or colder than a controlled room with a specific temperature and neutral meteorological conditions.
Once again, this exactly fits into our streaming active labeling setting.
The ``Weather'' dataset results are reported on the right of Figure \ref{sgd:fig:real_world}.
The baseline consists in predicting the apparent temperature as the real temperature.
We observe a transitory regime where the first 1,000 samples seem to be used to calibrate the weights $\alpha$.
During this regime, our estimate is too bad for the active strategy to make smarter queries than the ``random'' ones that have been calibrated on temperature statistics.
The main difference in the learning dynamic between the active and passive strategies is observed on the remaining 69,000 training samples.
The parameters were the same as the ``CalCOFI'' dataset but for $\gamma_0 = 10^{-2}$.

\end{document}